\documentclass[11pt]{article}

\usepackage[final]{acl}

\usepackage{times}
\usepackage{latexsym}

\usepackage[inkscapelatex=false]{svg}

\usepackage{booktabs}
\usepackage{multirow}
\usepackage{amsmath} 
\usepackage{amsfonts}
\usepackage{amssymb}
\usepackage{graphicx}
\usepackage{subcaption}
\usepackage{pgfplots}
\pgfplotsset{compat=1.18}

\usepackage{times}
\usepackage{latexsym}
\usepackage{booktabs}
\usepackage{color, soul}
\usepackage[T1]{fontenc}

\usepackage[utf8]{inputenc}
\usepackage[inkscapelatex=false]{svg}
\usepackage{microtype}

\usepackage{inconsolata}
\usepackage{multirow, makecell}

\usepackage{graphicx}
\usepackage{amsmath}

\usepackage{graphicx}
\usepackage{subcaption}

\sethlcolor{pink}

\usepackage[T1]{fontenc}

\usepackage[utf8]{inputenc}

\usepackage{microtype}

\usepackage{inconsolata}

\usepackage{graphicx}

\title{Additive Large Language Models for Semi-Structured Text}

\author{Karthikeyan K \\ Duke University \\ karthikeyan.k@duke.edu
        \AND
        Raghuveer Thirukovalluru \\ Duke University \\ raghuveer.thirukovalluru@duke.edu
        \And
        David Carlson\\ Duke University \\ david.carlson@duke.edu }
        
\newcommand{\rt}[1]{}

\begin{document}
\maketitle
\begin{abstract}
Large Language Models have advanced clinical text classification, but their opaque predictions remain a critical barrier to practical adoption in research and clinical settings where investigators and physicians need to understand which parts of a patient's record drive risk signals. To address this challenge, we introduce \textbf{CALM}, short for \textbf{Classification with Additive Large Language Models}, an interpretable framework for semi-structured text where inputs are composed of semantically meaningful components, such as sections of an admission note or question–answer fields from an intake form. CALM predicts outcomes as the additive sum of each component's contribution, making these contributions part of the forward computation itself and enabling faithful explanations at both the patient and population level. The additive structure also enables clear visualizations, such as component-level risk curves similar to those used in generalized additive models, making the learned relationships easier to inspect and communicate. Although CALM expects semi-structured inputs, many clinical documents already have this form, and similar structure can often be automatically extracted from free-text notes. CALM achieves performance comparable to conventional LLM classifiers while improving trust, supporting quality-assurance checks, and revealing clinically meaningful patterns during model development and auditing.

\end{abstract}

\section{Introduction}


The availability of large-scale patient data has accelerated the development of machine learning in healthcare, enabling models that detect patterns in complex clinical data for tasks such as {risk prediction} (e.g., ICU mortality)~\cite{che2017interpretable,mcnamara2016predicting}, {diagnostic support} (e.g., distinguishing cancer types)~\cite{wu2021breast}, and {prognosis estimation} (e.g., hospital readmission)\cite{kourou2015machine,zhang2023machine,diller2019machine}. These research applications demonstrate the potential of machine learning to improve clinical decision-making and outcomes, yet translation from research success to routine clinical adoption remains limited. Clinicians often rely instead on simple, rule-based scores such as the Framingham Risk Score for cardiovascular disease~\cite{mahmood2014framingham} or \texttt{APACHE}~\cite{larvin1989apache} and \texttt{SOFA}~\cite{ferreira2001serial,lambden2019sofa} for ICU mortality risk. Although less accurate than modern deep learning models, these tools remain popular because the contributing factors—such as heart rate, blood pressure, or lab values—are explicit and interpretable. Models that output only a “high-risk’’ label offer little value if users cannot see which factors shaped the prediction or by how much, underscoring the need for interpretability to make machine-learning outputs actionable and trustworthy in clinical settings.

Most inherently interpretable machine-learning approaches have been developed for structured, tabular data, where each patient is represented by a fixed-length vector of numeric or categorical variables (e.g., vitals, lab results, demographics). These tabular models include sparse linear scoring systems, rule-based models, generalized additive models, and constrained trees or ensembles \cite{ustun2016slim,ustun2017riskslim,angelino2018learning,lakkaraju2016interpretable,lou2013accurate,caruana2015intelligible,nori2019interpretml,freund1997decision,Speybroeck2012ClassificationAR}. These methods work well on standardized features, but tabular models may miss important nuances of clinical care and vary across institutions, which limits performance and transferability. In contrast, text-based classifiers operate directly on unstructured or semi-structured clinical narratives such as admission notes, progress notes, or discharge summaries. Clinical text can capture richer observations about a patient’s status (e.g., evolving symptoms, clinician impressions, or social context often missing from structured fields). Recent advances in LLMs position text as a promising foundation for inherently interpretable methods that can generate meaningful insights and support real-world decision-making.

Unlike tabular models, text-based classifiers in healthcare are typically opaque, and interpretability has often been sought through post-hoc explanations of black-box models. Common approaches include {local surrogate models} \cite{ribeiro2016lime}, {saliency- or attribution-based techniques} \cite{lundberg2017shap,sundararajan2017axiomatic,zeiler2014visualizing,fong2017meaningful}, {attention-based rationales} \cite{bahdanau2015neural,jain2019attention}, and {influence-function methods} that trace predictions to training examples \cite{koh2017understanding,pruthi2020estimating,wen2024languagemodelmeetsprototypes}. However, these explanations often fail to capture true model reasoning: saliency maps can be insensitive to parameter or data changes \cite{adebayo2018sanity}, attention weights may not align with feature importance \cite{jain2019attention}, and removing tokens flagged as important frequently leaves predictions unchanged \cite{hooker2019benchmark}. Such limitations undermine trust in clinical applications and highlight that opaque models with approximate explanations are not a substitute for inherently interpretable models\cite{rudin2019stop}. These challenges motivate the development of text-native, inherently interpretable methods that can provide faithful insights into model behavior.

Despite substantial progress on interpretable models for tabular data, text-based prediction still relies largely on post-hoc explanations of black-box models, which remain unreliable for high-stakes settings. To address this gap, we propose \textbf{CALM} (Classification with Additive Large Language Models), an inherently interpretable framework for predictive modeling with text. CALM is designed for semi-structured inputs, where each document—such as a patient note—can be decomposed into meaningful components like sections of clinical documents, question–answer fields, or survey responses. Recent tools such as Clinstructor \cite{clinstructor} can automatically transform free-text notes into this semi-structured form. By enforcing an additive decomposition across components, CALM produces faithful component-level contributions that make predictions transparent and readily visualized.


We evaluate CALM on three real-world clinical outcome tasks. Across all tasks, CALM performs competitively with black-box LLM classifiers, achieving comparable AUC-PR while offering inherent interpretability at both the patient- and population-level. Variants that capture pairwise interactions (CALM$^2$) or distill knowledge from fine-tuned teachers further close the already small performance gap. These results demonstrate that additive LLM-based models can provide transparency without sacrificing performance, offering a practical framework for building trustworthy machine-learning systems in healthcare.

\section{Background and Related Works}

\paragraph{Neural Additive Models (NAMs)}~\cite{agarwal2021neural,hastie2017generalized}are inherently interpretable architectures that learn a distinct neural network for each individual input feature. The final prediction is computed as the bias term plus the sum (or mean) of the logits from these feature-specific subnetworks, allowing users to directly observe each feature’s contribution to the outcome. Several variants of NAMs such as NA2M, GAMI-Net have also been proposed that incorporate pairwise interactions~\cite{yang2021gami,chang2021node,xu2023sparse,peroni2022extending,enouen2022sparse,chang2021node}. Unlike NAMs and its variants, where the input to each neural model is a scalar feature, CALM operates over textual inputs—each feature corresponds to a piece of text such as a section from a patient’s notes or a question–answer pair and uses shared LLMs as  feature-specific subnetworks.

\paragraph{ClinStructor}~\cite{clinstructor}is a three-stage pipeline that uses Large Language Models (LLMs) to convert unstructured clinical notes into structured question-answer pairs for predictive modeling. First, it identifies the most relevant clinical questions based on the patient notes (Feature Identification). Next, it extracts answers to these questions from each patient's record (Feature Extraction). Finally, it fine-tunes a language model on the concatenated question-answer-pairs. We follow the the same process as ClinStructor's question-answer creation and use it as one of the dataset for our experiments.

\paragraph{Post-hoc Explanation Methods:}Common approaches include feature-attribution techniques such as SHAP~\cite{lundberg2017shap,mosca2022shap} and Integrated Gradients~\cite{sundararajan2017axiomatic}, local surrogate models like LIME~\cite{ribeiro2016lime}, attention visualization methods~\cite{vig2019multiscale,yeh2023attentionviz}, and similarity-based approaches such as K-Nearest Neighbors~\cite{papernot2018deep} and Influence Functions~\cite{koh2017understanding,pruthi2020estimating}, which relate predictions to influential or similar examples from the training data. However, such post-hoc explanation may not be faithful or reliable explanations and may not accurately reflect a model’s true reasoning process~\cite{rudin2019stop,basu2020influence,karthikeyan2021revisiting,slack2020fooling,dasgupta2022framework,barr2023disagreement,miro2024assessing,bhalla2023discriminative}. Unlike post-hoc explanation methods, CALM is an inherently interpretable architectures, that provides both local and global interpretability.

\section{CALM: Classification with Additive Large Language Models}

We propose CALM, a general and interpretable framework for text classification. The input to CALM, denoted as $x = {x_1, x_2, \dots, x_M}$, consists of a collection of $M$ textual components, where each component $x_i$ represents a semantically meaningful text segment—such as a section of a clinical note, a question–answer pair, or a response from a survey form. While CALM requires semi-structured text as input, it can also be applied to unstructured text. In such cases, we can first transform the text into a semi-structured form suitable for CALM by following the segmentation procedures introduced in ClinStructor~\cite{clinstructor}. 

Before introducing the CALM architecture, we first outline the standard fine-tuned LLM classifier, which serves as our baseline.

\subsection{Baseline: Standard LLM Fine-tuning}

In the standard text classification approach, all input components are concatenated into a single sequence $\tilde x = x_ix_2..x_M$. This sequence is then passed through a Large Language Model (LLM) backbone $F_T(\cdot; w_T)$, followed by a classification head $F_{\text{last}}(\cdot; W_{\text{last}})$. The logits $z \in \mathbb{R}^C$ and probabilities $p_c$ for $C$ classes are computed as
\begin{align*}
\textstyle
    z & = f_{\text{last}}\!\big(f_T(\tilde x; w_T), w_{\text{last}}\big),\\
    \textstyle
    p_c & = \textstyle\frac{\exp(z_c)}{\sum_{j=1}^C \exp(z_j)}, \quad c = 1,\dots,C.
\end{align*}
The model is trained by minimizing a classification loss $\mathcal{L}(y_{true},p)$, such as cross-entropy. While such fine-tuning typically achieves strong predictive performance, its critical limitation is a complete lack of interpretability.

\subsection{CALM Architecture}

The $\text{CALM}$ architecture modifies the standard classifier by enforcing an additive structure on the final prediction, ensuring component-level interpretability. Each input component $x_i$ is first independently encoded by the shared LLM backbone $F_T(\cdot; w_T)$. The resulting representation $h_i$ is then mapped to class logits $\ell_i \in \mathbb{R}^C$ by a component-specific classification head $F^i_{\text{last}}(\cdot; W^i_{\text{last}})$:
\begin{align*}
    h_i  = f_T(x_i; w_T); \quad 
    \ell_i = f^i_{\text{last}}(h_i; w^i_{\text{last}}).
\end{align*}
The overall logit vector $z \in \mathbb{R}^C$ is the average of the component contributions, with a learned bias $b$:
\begin{equation*}
\textstyle
    z = \frac{1}{M}\sum_{i=1}^M \ell_i + b.
\end{equation*}
Predicted class probabilities $p_c$ are then calculated using the $\text{softmax}$ function over $z$. The model is trained using the same loss $\mathcal{L}(y_{true},p)$ as the standard fine-tuned LLM classifier. This additive formulation inherently exposes the influence of every component directly at the logit level, as $\ell_i$ depends only on $x_i$ and naturally represents its contribution to the final prediction.

\paragraph{Interpretability}
The additive structure provides intrinsic patient-level interpretability by outputting the contribution $\ell_i$ for each component $x_i$. These contributions quantify the magnitude and direction of influence each component has on the final prediction, allowing the decision to be directly traced to specific input evidence. Furthermore, aggregating $\ell_i$ across a dataset provides model-level interpretability, quantifying the global importance of different input components across the population. A detailed analysis and discussion of $\text{CALM}$ interpretability is provided in Section~\ref{sec:Interpretability}.

\paragraph{Time Complexity}
Let $L_i := |x_i|$, $L_{\max} := \max_i L_i$, and $L_{\mathrm{tot}} := \sum_{i=1}^{M} L_i$. A standard fine-tuned classifier processes the concatenated sequence in $O(L_{\mathrm{tot}}^2)$ time. In contrast, $\text{CALM}$ processes each component independently, resulting in a theoretical total cost of $\sum_{i=1}^M O(L_i^2)$. Since $\sum_{i=1}^M L_i^2 < L_{\mathrm{tot}}^2 $ for $M>1$, $\text{CALM}$ offers significant efficiency improvements. For $M$ equal-length components, $\text{CALM}$ achieves an $M$-fold speedup ($M L^2$ vs. $M^2 L^2$). Thus, $\text{CALM}$ is not only interpretable but can also be significantly more efficient than direct fine-tuning, provided component lengths are not highly imbalanced. In practice, components are often batched by padding to $L_{\max}$, yielding a practical cost of $O(M L_{\max}^2)$, which can occasionally be larger than $O(L_{\mathrm{tot}}^2)$. To mitigate this, we discuss an alternative packed implementation in the Appendix that is mathematically equivalent and has an overall cost of at most $O(L_{\mathrm{tot}}^2)$.

\subsection{\texorpdfstring{CALM\textsuperscript{2}}: Modeling Pairwise Interactions}\label{sec:calm2}

While the CALM architecture enforces independent component contributions to the final prediction, considering components jointly can reveal a stronger signal than when viewed in isolation. For instance, certain lab values may appear normal on their own, yet become clinically significant when interpreted alongside a patient's age from another segment. To effectively capture such interactions, we propose CALM$^2$, an extension of CALM that incorporates explicit pairwise interaction terms. Pairwise interactions can be incorporated in two principal ways.

\paragraph{(i) Text-level interactions.}
This approach processes the $M$ individual components along with all $\binom{M}{2}$ pairwise concatenations $[x_i; x_j]$. Each combined sequence is passed through the LLM backbone $f(\cdot; w_T)$ and a pair-specific classification head $g^{ij}(\cdot; w^{ij})$. However, the overall computation cost is prohibitively high, as this requires $O(M^2)$ forward passes over longer sequences, making the approach impractical for moderate or large $M$.

\paragraph{(ii) Embedding-level interactions.}
A more efficient alternative computes the embeddings of each component $h_i = f(x_i; w_T)$ and then defines interaction terms $\ell_{ij}$ in the hidden space using a lightweight interaction function $g^{ij}$, where $\ell_{ij} = g^{ij}(h_i, h_j; w^{ij})$. The final logits are
\begin{align*}
\textstyle
    z &\textstyle = \frac{1-\beta}{M}\sum_{i=1}^M \ell_i + \frac{\beta}{\binom{M}{2}} \sum_{1 \leq i < j \leq M} \ell_{ij}.
\end{align*}
Here, $\beta$ is an inductive weight, representing the prior belief on how much to rely on the interaction logits. This approach reuses the CALM computations and adds pairwise terms through lightweight operations, maintaining near CALM complexity while providing greater expressivity.

\paragraph{Choice of $g^{ij}$.}
We instantiate $g^{ij}$ using a low-rank bilinear interaction model. Each embedding $h_i$ is first projected into an $R$-dimensional latent space ($\tilde h_i = U_i h_i$, $\hat h_j = U_j h_j$), and their interaction is captured by an elementwise (Hadamard) product, followed by a pair-specific classifier head:
\begin{align*}
    \ell_{ij} &= w_{\mathrm{out}}^{ij} (\tilde h_i \odot \hat h_j), \quad w_{\mathrm{out}}^{ij} \in \mathbb{R}^{C \times R}.
\end{align*}
By choosing a small rank $R$ (e.g., $8$ or $16$), the additional parameters and computation introduced by the quadratic number of pairs remain limited, with an overall complexity of $O(M^2 R)$. Note that CALM$^2$ remains highly interpretable, as each interaction term $\ell_{ij}$ depends only on $(x_i, x_j)$. This allows us to visualize $\ell_{ij}$ as a function of $(x_i, x_j)$ using heatmaps or 3D plots, analogous to the pairwise analysis in $\text{GA}^2\text{M}$s.

\begin{table*}[t!]
\centering
\begin{tabular}{llcccccc}
\toprule
\multirow{2}{*}{Dataset} & \multirow{2}{*}{Model} &
\multicolumn{3}{c}{Finetune} & \multicolumn{3}{c}{CALM} \\
\cmidrule(lr){3-5} \cmidrule(lr){6-8}
& & AUC-PR & F1 & AUC-ROC & AUC-PR & F1 & AUC-ROC \\

\toprule
\multirow{8}{*}{ClinStructor} 
 &  Qwen3-0.6B-Base & 0.43 & 0.449 & 0.839 &  0.418 & 0.441 & 0.825 \\
 &  Qwen3-1.7B-Base &  0.458 & 0.464 & 0.855 & 0.427 & 0.439 & 0.826 \\
 &  Qwen3-8B-Base & 0.467 & 0.474 & 0.857 &  0.451 & 0.466 & 0.842 \\
 & Phi-3.5-mini-instruct & 0.454 & 0.477 & 0.857  & 0.450 & 0.455 & 0.847 \\
 &  MediPhi          &  0.465 & 0.473 & 0.854 &  0.442 & 0.465 & 0.841 \\
 &  gemma-3-4b-pt       &   0.469 & 0.47 & 0.856 & 0.428 & 0.441 & 0.832 \\
 &  medgemma-4b-pt      &  0.448 & 0.467 & 0.852 & 0.440 & 0.447 & 0.842 \\
\toprule
\multirow{8}{*}{MIMIC-III} 
 &  Qwen3-0.6B-Base & 0.457 & 0.471 & 0.85 & 0.448 & 0.457 & 0.841 \\
 &  Qwen3-1.7B-Base &  0.494 & 0.481 & 0.859 & 0.470 & 0.469 & 0.849 \\
 &  Qwen3-8B-Base & 0.532 & 0.506 & 0.874  & 0.496 & 0.5 & 0.862 \\
 & Phi-3.5-mini-instruct &  0.522 & 0.506 & 0.87  & 0.482 & 0.477 & 0.854
\\
 &  MediPhi          &  0.508 & 0.486 & 0.866  & 0.455 & 0.475 & 0.851 \\
 &  gemma-3-4b-pt       &  0.51 & 0.49 & 0.868  & 0.481 & 0.487 & 0.862
 \\
 &  medgemma-4b-pt      &  0.514 & 0.491 & 0.866  & 0.479 & 0.478 & 0.859
 \\
\toprule
\multirow{6}{*}{LCD} 
 & Qwen3-0.6B-Base & 0.251 & 0.315 & 0.872  &  0.262 & 0.292 & 0.851 \\
 & Qwen3-1.7B-Base  &  0.258 & 0.332 & 0.877 & 0.268 & 0.312 & 0.867  \\
 &  gemma-3-4b-pt         & 0.267 & 0.33 & 0.865 &  0.276 & 0.336 & 0.867\\
 &  medgemma-4b-pt      &  0.249 & 0.311 & 0.858 & 0.251 & 0.307 & 0.863 \\
\bottomrule
\end{tabular}
\caption{\textbf{Comparison of CALM and Black-box LLM Finetuning} }\label{tab:calm}
\end{table*}

\subsection{CALM-Distill: Enhancing CALM with Knowledge Distillation}\label{sec:calmdistill}

The CALM architecture achieves efficient and inherently interpretable classification via its additive structure. To further enhance its predictive capabilities, we introduce CALM-Distill, an extension that transfers knowledge from a strong, fully fine-tuned black-box teacher LLM while preserving CALM’s interpretable architecture.

The teacher model produces logits $z^{(T)}$, and the CALM student produces additive logits $z^{(S)} = \frac{1}{M} \sum_{i=1}^M \ell_i + b$. CALM-Distill training minimizes a combined objective function $\mathcal{L}$ that is a convex combination of the ground-truth cross-entropy loss ($\mathcal{L}_{\mathrm{CE}}$) and the knowledge distillation loss ($\mathcal{L}_{\mathrm{KD}}$).

The distillation loss $\mathcal{L}_{\mathrm{KD}}$ is the $T^2$-scaled Kullback-Leibler (KL) divergence between the student and teacher distributions, $q^{(S)}$ and $q^{(T)}$, which are calculated from the logits and temperature $T$ as follows:
\begin{align*}
q^{(T)} &= \mathrm{softmax}\!\left(\tfrac{z^{(T)}}{T}\right), \quad q^{(S)} = \mathrm{softmax}\!\left(\tfrac{z^{(S)}}{T}\right)\\
\mathcal{L}_{\mathrm{KD}} &= T^2 \, \mathrm{KL}\!\left(q^{(T)} \,\|\, q^{(S)}\right) \\
\mathcal{L} &= (1-\alpha)\,\mathcal{L}_{\mathrm{CE}} + \alpha\,\mathcal{L}_{\mathrm{KD}},
\end{align*}
where the hyperparameter $\alpha \in [0,1]$ balances ground-truth supervision against the teacher’s guidance. This process effectively transfers the dark knowledge -- the relative probabilities and uncertainties encoded in the teacher’s soft targets -- to the student. CALM-Distill leverages teacher's knowledge without modifying its architecture, thereby retaining full interpretability.

\section{Experiments and Results}

\subsection{Datasets}
We evaluate $\text{CALM}$ and its extensions on three real-world datasets derived from Intensive Care Unit (ICU) patient notes: (1) MIMIC Admission Notes, (2) ClinStructor, and (3) the Long Clinical Document (LCD) benchmark. All three datasets exhibit significant class imbalance. To address this, we subsample the training data by retaining all positive examples and randomly sampling an equal number of negative examples. The original class distributions are strictly maintained for the validation and test sets.

\paragraph{MIMIC Admission Notes}~\cite{van-aken-etal-2021-clinical}
This dataset is constructed from $\text{MIMIC-III}$ discharge summaries by retaining only the textual sections typically recorded upon a patient's $\text{ICU}$ arrival (e.g., Chief Complaint, Medical History, and Physical Exam). We preprocess and segment these into eight distinct sections, where each section serves as an input component, $x_i$, for the $\text{CALM}$ framework. The objective is to predict in-hospital mortality: whether a patient dies during the $\text{ICU}$ stay or is discharged alive. After subsampling the training data, the final dataset comprises 7,027 training examples (3,506 deceased, 3,521 alive), 4,882 validation examples (514 deceased, 4,368 alive), and 9,773 test examples (1,020 deceased, 8,753 alive).

\begin{table*}[t!]
\centering
\begin{tabular}{lcccccc}
\toprule
\multirow{2}{*}{Model} & \multicolumn{4}{c}{CALM$^2$ (AUC-PR)} & \multirow{2}{*}{Finetune} & \multirow{2}{*}{CALM} \\
\cmidrule(lr){2-5}
& R=8, $\beta$=0.1 & R=8,$\beta$=0.5 & R=16, $\beta$=0.1 & R=16, $\beta$=0.5 & & \\
\midrule
Qwen3-0.6B-Base         & 0.447 & 0.445 & 0.437 & \textbf{0.464} &\underline{0.457} & 0.448  \\
Qwen3-1.7B-Base         & 0.476 & \underline{0.486} & 0.475 & 0.464 & \textbf{0.494} & 0.470 \\
Qwen3-8B-Base           & \underline{0.504} & 0.482 & 0.485 & 0.503 & 
\textbf{0.532} & 0.496 \\
Phi-3.5-mini-instruct   & 0.480 & 0.462 & 0.473 & 0.476 & \textbf{0.522} & \underline{0.482} \\
gemma-3-4b-pt           & 0.487 & 0.500 & 0.487 & \underline{0.492} & \textbf{0.510} & 0.481\\
\bottomrule
\end{tabular}
\caption{\textbf{Comparison of CALM$^2$ and CALM:} We report AUC-PR on the MIMIC admission notes dataset. CALM$^2$ outperforms CALM and further narrows the gap to black-box fine-tuning.}\label{tab:calm2}
\end{table*}

\begin{table*}[t!]
\centering
\begin{tabular}{lcccccc}
\toprule
\multirow{2}{*}{Model} & \multicolumn{3}{c}{CALM-Distill (AUC-PR)} & \multirow{2}{*}{Finetune/Teacher} & \multirow{2}{*}{CALM} \\
\cmidrule(lr){2-4}
& $\alpha=0.2$ & $\alpha=0.4$ & $\alpha=0.6$ &  & \\
\midrule
Qwen3-0.6B-Base   & \underline{0.454} &  0.445 &  0.45 & \textbf{0.457} & 0.448 \\
Qwen3-1.7B-Base   &  0.479 &  \underline{0.486} &  0.477 & \textbf{0.494} & 0.470   \\
Qwen3-8B-Base   &  0.512 &  \underline{0.517} &  \underline{0.517} &   \textbf{0.532} & 0.496   \\
Phi-3.5-mini-instruct   & 0.486 &  0.495 & \underline{0.503} &  \textbf{0.522} & 0.482  \\
gemma-3-4b-pt & 0.505 &  0.504 &  \textbf{0.524} & 
\underline{0.510} & 0.481\\
\bottomrule
\end{tabular}
\caption{\textbf{Comparison of CALM-Distill and CALM:} We report AUC-PR on the MIMIC admission notes dataset. CALM-Distill consistently outperforms CALM and further narrows the gap to black-box finetuning.}\label{tab:distill}
\end{table*}

\paragraph{ClinStructor}~\cite{clinstructor} is a structured representation of clinical notes derived from MIMIC Admission Notes. To create ClinStructor data, we follow the procedures in~\cite{clinstructor}; however, instead of Llama~3.3 and Qwen~2.5, we use Gemini-flash-2.5\footnote{We use Gemini through Vertex AI, which complies with the MIMIC data use agreement.}. The task setup and the number of examples are identical to the MIMIC Admission Notes dataset. Unlike MIMIC, ClinStructor comprises of 50 fine-grained question–answer pairs, each corresponding to one CALM component; see the appendix for the full list.

\paragraph{Long Clinical Document (LCD) Benchmark}~\cite{yoon2025lcd} consists of long discharge notes from the MIMIC-IV database for patients admitted to later discharged from ICU. Similar to MIMIC, we pre-process and split each note into 22 sections. The task is to predict 30-day out-of-hospital mortality. The final dataset includes 2,534 training examples (1,267 deceased and 1,267 alive), 7,505 validation examples (302 deceased and 7,203 alive), and 7,568 test examples (261 deceased and 7,307 alive).

\subsection{Experiment Setup}

\paragraph{Models:}We experimented with seven open-source LLMs—Qwen 3 (0.6B, 1.7B, 8B)~\cite{yang2025qwen3}, Phi-3.5-mini, MediPhi~\cite{corbeil2025modular}, Gemma-3 (4B)~\cite{team2025gemma}, and MedGemma-3 (4B)~\cite{sellergren2025medgemma}—spanning diverse parameter sizes, model families, and domains, including both general-purpose and medical LLMs.

\paragraph{Training Hyperparameters:} We ensure fair comparison across all experiments -- regular fine-tuning, CALM, and its variants -- by using LoRA fine-tuning for 5 epochs with batch size 1, gradient accumulation 16, LoRA dropout 0.05, and the AdamW optimizer. We tune learning rate (1e-4, 2e-4, 5e-4), LoRA rank (8, 16), and scaling factor (16, 32) across eight settings, applied consistently in all experiments (see Appendix). The best model is selected using validation data, and its performance is reported on the test set.

\paragraph{Metrics.} For all experiments, we report AUC-PR, F1, and AUC-ROC. AUC-PR is used for early stopping and for model selection.

\subsection{Results}

\begin{figure*}[t!]
    \centering
    \begin{subfigure}[b]{0.45\textwidth}
        \centering
        \includegraphics[width=\linewidth]{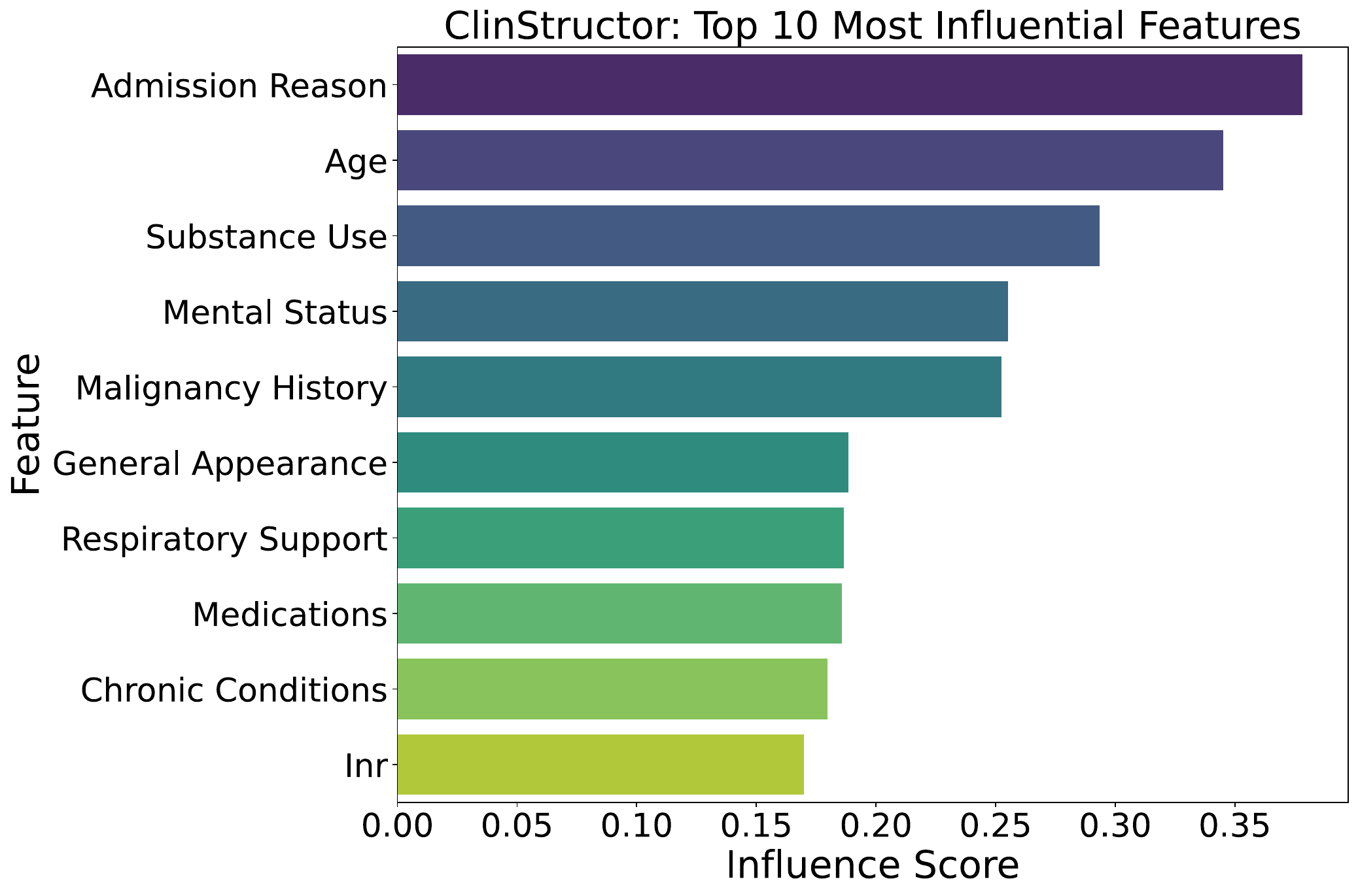}
        \caption{Influential features of ClinStructor}
        \label{fig:feature_importance_clinstructor}
    \end{subfigure}
    \hfill
    \begin{subfigure}[b]{0.49\textwidth}
        \centering
        \includegraphics[width=\linewidth]{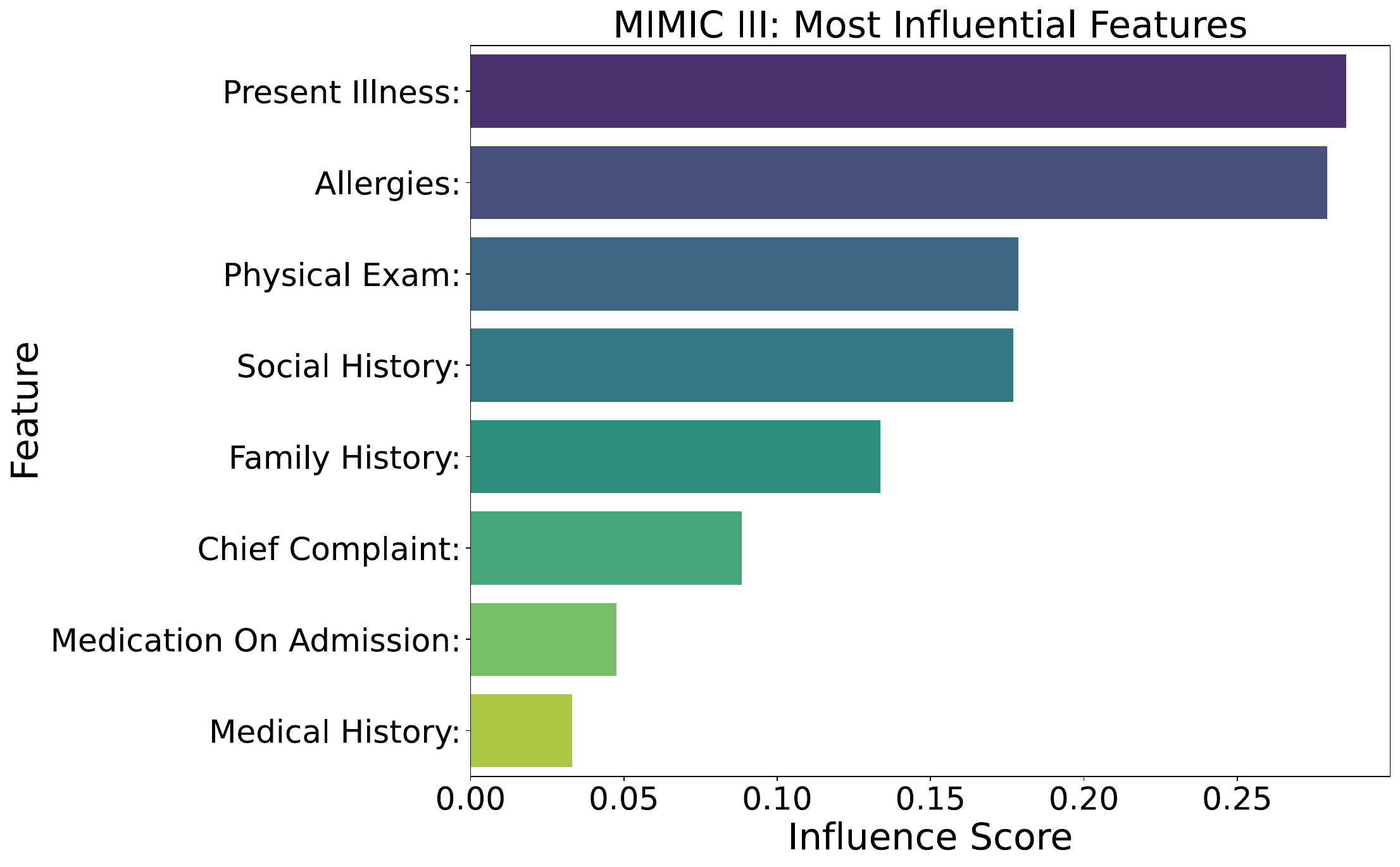}
        \caption{Influential features of MIMIC Admission Notes}
        \label{fig:mimic_modellevel}
    \end{subfigure}
    \caption{\textbf{Influential features:} The influence score is defined as the absolute risk score averaged across the population. The figure shows the top influential features and their scores from Clinstructor and MIMIC, respectively.}
    \label{fig:feature_importance}
\end{figure*}

\begin{figure*}[t!]
    \centering
    \begin{subfigure}[b]{0.24\textwidth}
        \centering
        \includegraphics[width=\linewidth]{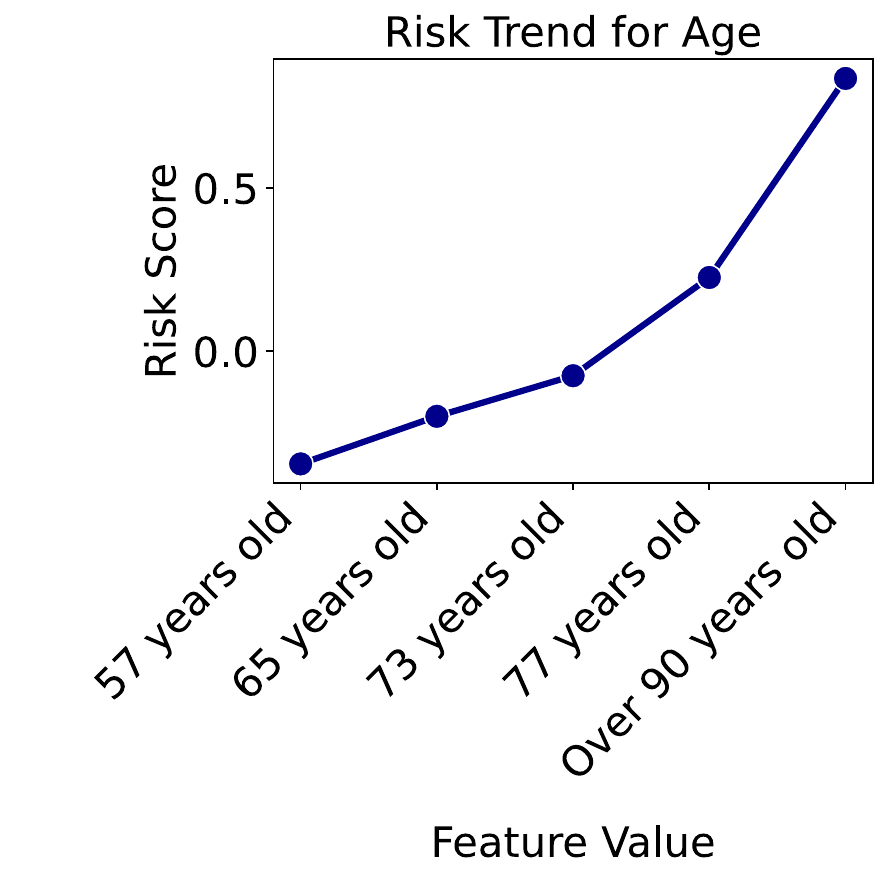}
        \caption{Age}
    \end{subfigure}
    \hfill
    \begin{subfigure}[b]{0.24\textwidth}
        \centering
        \includegraphics[width=\linewidth]{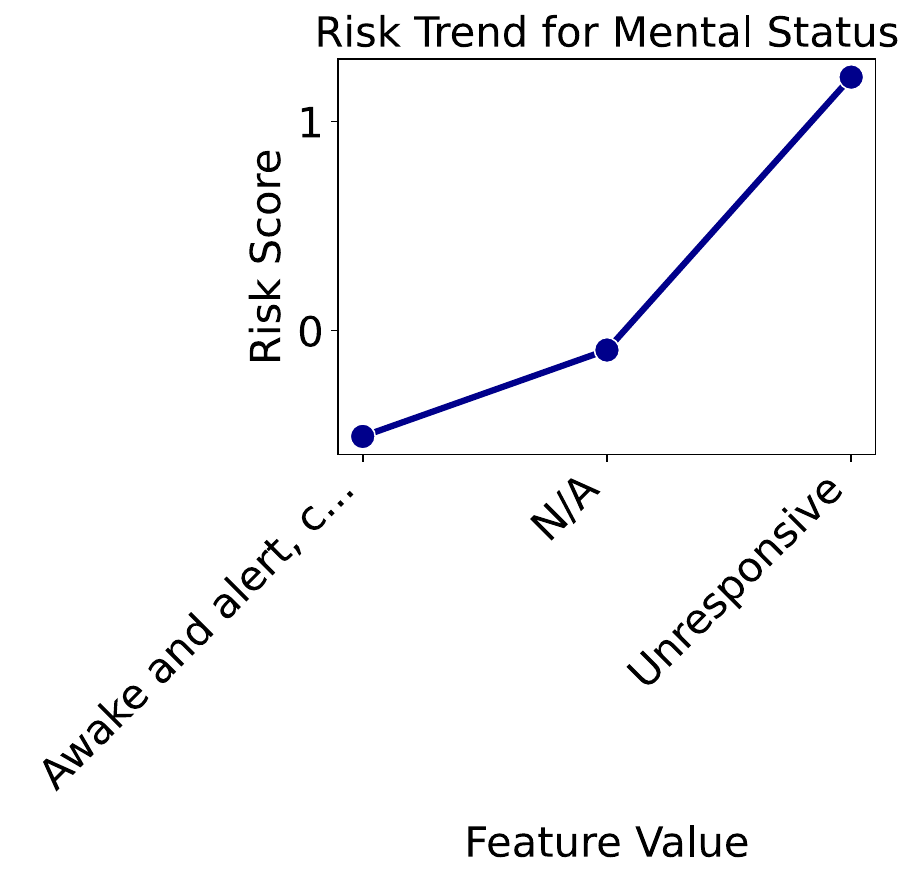}
        \caption{Mental Status}
    \end{subfigure}
    \hfill
    \begin{subfigure}[b]{0.24\textwidth}
        \centering
        \includegraphics[width=\linewidth]{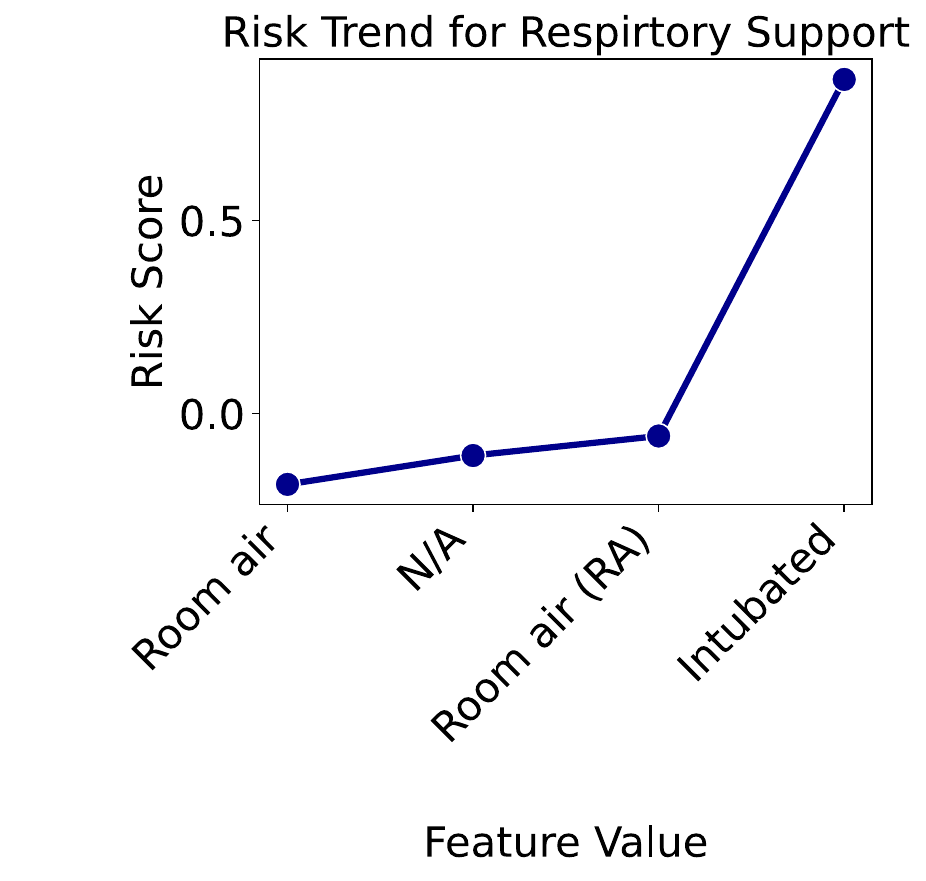}
        \caption{Respiratory Support}
    \end{subfigure}
    \hfill
    \begin{subfigure}[b]{0.24\textwidth}
        \centering
        \includegraphics[width=\linewidth]{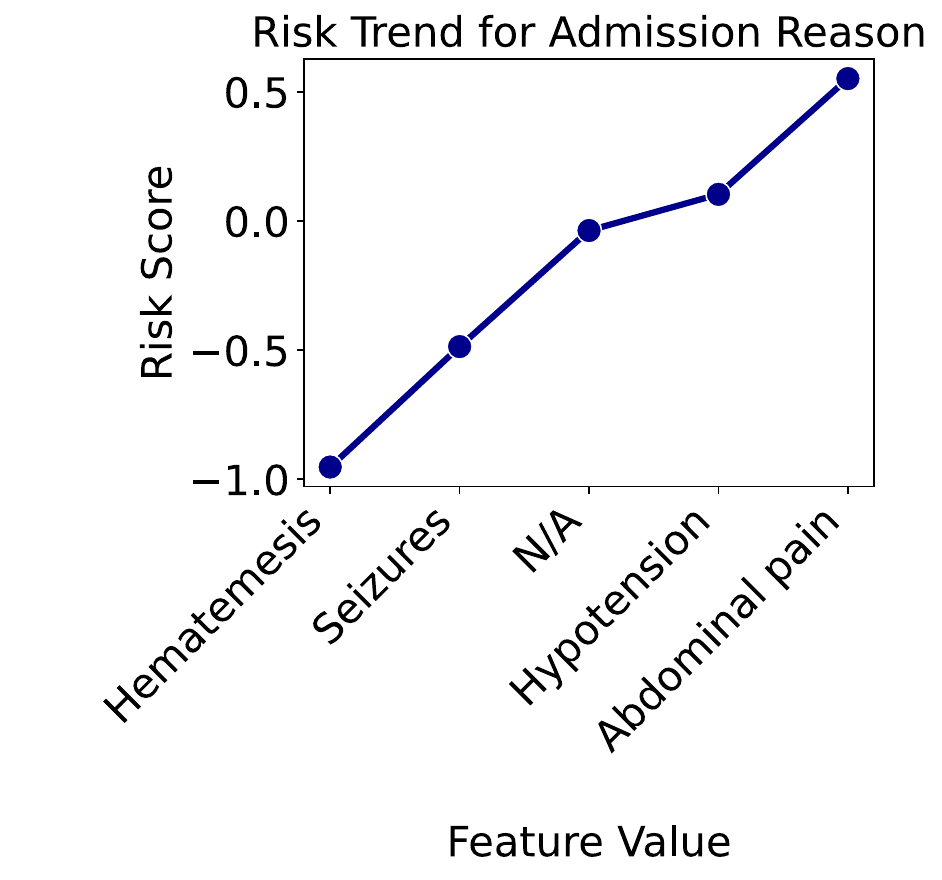}
        \caption{Admission Reason}
    \end{subfigure}
    \caption{\textbf{Feature values and corresponding risk scores:} We select the top 20 most frequent feature values and, among them, choose 5 values—spanning the range of logit scores—to visualize feature logit function $\ell_i(x_i)$.}
        \label{fig:RiskTrends}
\end{figure*}

\subsubsection{Black-box LLM Vs. CALM Finetuning:} 

Across three datasets and seven language models, we compare black-box LLM finetuning with our proposed CALM approach. From Table~\ref{tab:calm}, CALM performs competitively with finetuning baseline. performance drops by $0.02$ and $0.03$ AUC-PR on ClinStructor and MIMIC dataset respectively. On the LCD data, CALM performs on par with finetuning. Similar to NAMs, minor decrease in performance is an expected trade-off for interpretability. Nonetheless, the trade-off is worthwhile, as CALM’s interpretability makes the model far more actionable than black-box predictions.

We evaluated the LCD dataset on only four models, as its long clinical notes often exceed the token limits of \textit{Phi-3.5-mini-instruct} and \textit{MediPhi}. The large context length also makes finetuning with Qwen 8B and even Gemma-3 or MedGemma 4B challenging. To address this, we used an alternate implementation described in the appendix, which is mathematically identical to CALM.

\noindent \textbf{Takeaway:} CALM performs competitively with black-box finetuning but, with its interpretability, CALM predictions that are more actionable and trustworthy.

\subsubsection{\texorpdfstring{CALM\textsuperscript{2}} Performance Improvements} The reduced performance of CALM arises from its strict additive constraint, which disallows interactions between components. To assess whether pairwise interactions enhance predictive power, we propose CALM$^2$ and compare it with CALM and black-box finetuning using five representative models on the MIMIC admission dataset.

As described in \S\ref{sec:calm2}, CALM$^2$ introduces two additional hyperparameters: the inductive weight of interaction logits $\beta$ and the rank $R$ of the low-rank bilinear projection matrices. We test $R \in {8, 16}$ and $\beta \in {0.1, 0.5}$. Table~\ref{tab:calm2} shows that CALM$^2$ outperforms CALM and further narrows the gap with black-box finetuning.
\noindent 

\noindent \textbf{Takeaway: }CALM$^2$ improves performance while still maintaining full interpretability.

\subsubsection{Improvements with CALM-Distill}As described in \S~\ref{sec:calmdistill}, we enhance CALM by distilling \textit{dark knowledge} from a teacher model. For fair comparison, CALM, CALM-Distill, and the teacher (black-box finetuned) model share the same LLM backbone and training data. CALM-Distill introduces two additional hyperparameters—temperature and the trade-off parameter $\alpha$ (weight of the KL loss). We fix the temperature at 2 and test $\alpha \in {0.2, 0.4, 0.6}$. Table~\ref{tab:distill} shows that CALM-Distill consistently outperforms CALM and further reduces the gap with black-box finetuning. Since only the optimization objective changes, CALM-Distill remains fully interpretable. 

\noindent \textbf{Takeaway: }CALM-Distill consistently improves over CALM while preserving full interpretability.

\begin{figure*}[t!]
    \centering
    \begin{subfigure}[b]{0.45\textwidth}
        \centering
        \includegraphics[width=\linewidth]{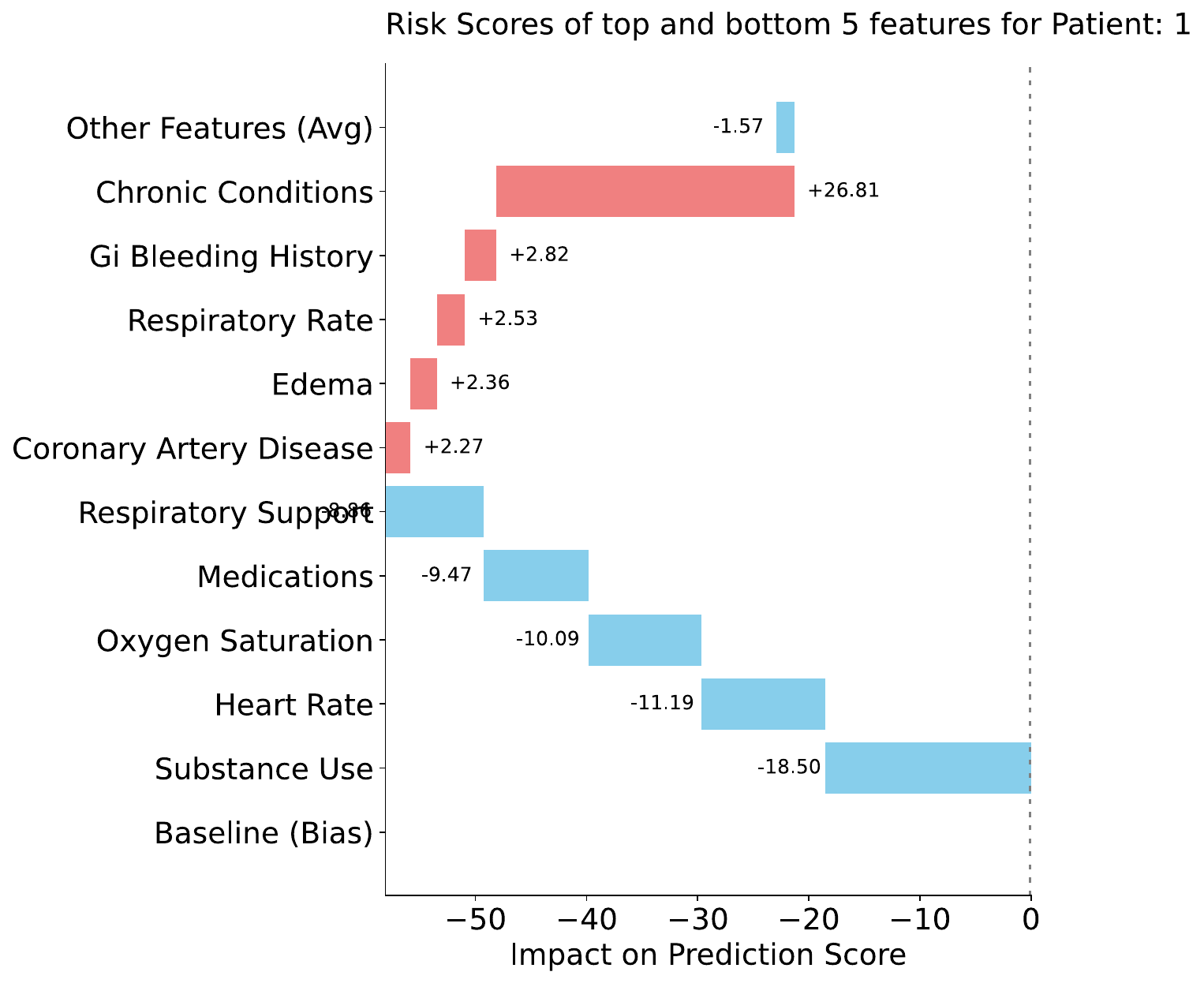}
        \caption{Patient 1: Risk scores }
        \label{fig:plot1}
    \end{subfigure}
    \hfill
    \begin{subfigure}[b]{0.45\textwidth}
        \centering
        \includegraphics[width=\linewidth]{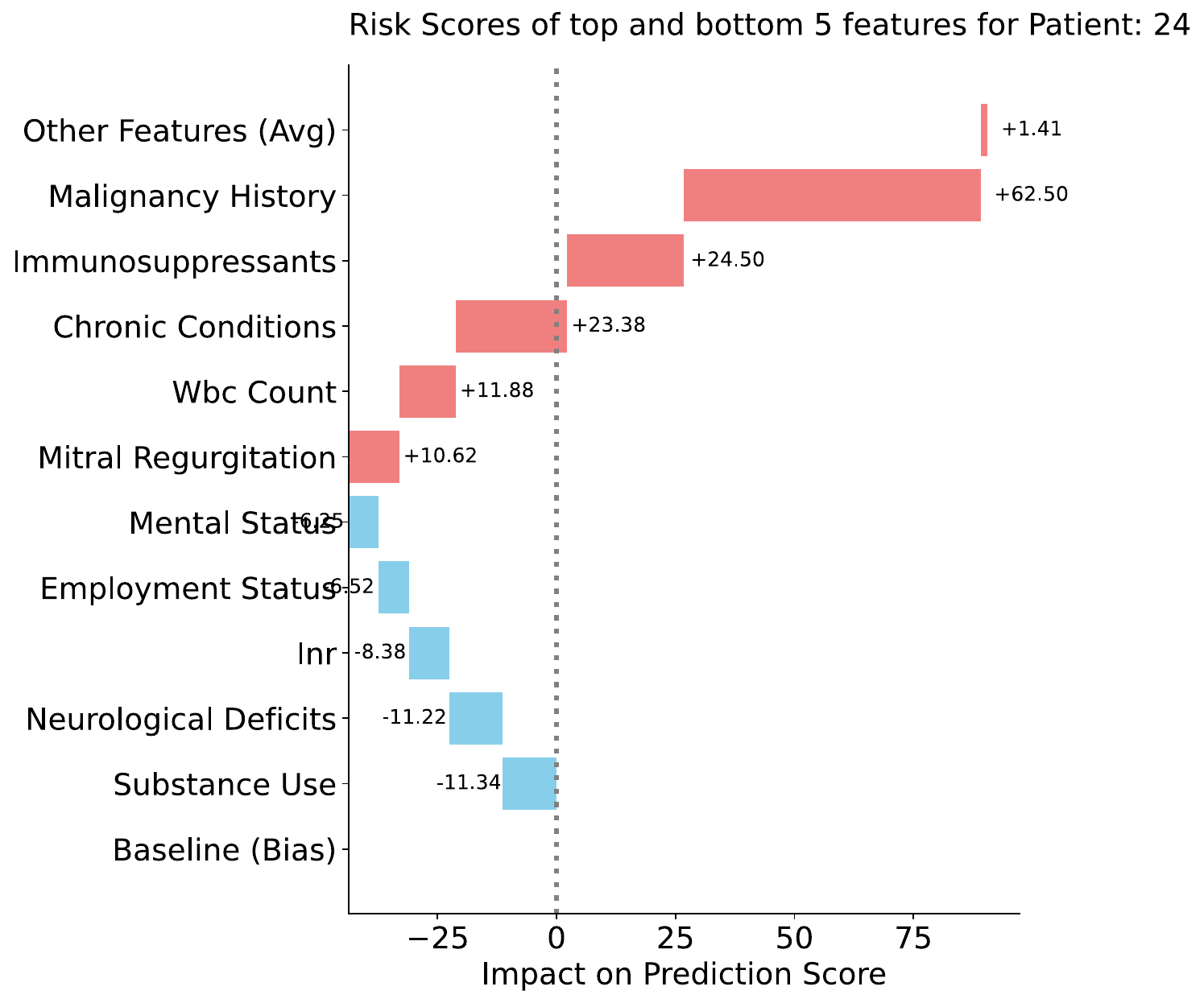}
        \caption{Patient 24: Risk scores }
        \label{fig:plot24}
    \end{subfigure}
\caption{\textbf{Patient-level interpretability:} We plot the top 5 and bottom 5 risk score features and their corresponding scores for two patients from ClinStructor—one who died and one who was discharged alive.}\label{fig:patientinterpretability}
\end{figure*}

\section{Global and Local Interpretability}\label{sec:Interpretability}

CALM provides inherent interpretability at both global (model-level) and local (patient-level) scales. In this section, we illustrate both.

\subsection{Global Interpretability}

\paragraph{Feature Importance Scores: } 


Feature importance scores quantify, across the population, how each input component contributes to the model’s decision. For each patient $n$ and input component $i$, CALM produces logit scores $[\ell^{(0)}_{i,n}, \ell^{(1)}_{i,n}]$. The difference $\ell^{(1)}_{i,n} - \ell^{(0)}_{i,n}$ represents the feature’s contribution toward label~1; a large positive value ($\gg 0$) indicates strong support for label~1, while a large negative value ($\ll 0$) indicates strong support for label~0. The magnitude $|\ell^{(1)}_{i,n} - \ell^{(0)}_{i,n}|$ reflects the strength of influence, and its sign indicates direction. The overall influence of a feature $i$ is computed as the average magnitude of this difference across all patients:

\begin{align*}
    &\text{Influence} \textstyle(i)=  \frac{1}{N}\sum_{n=1}^N |\ell_{i,n} ^{(1)}-\ell_{i,n}^{(0)} | 
\end{align*}


Figure~\ref{fig:feature_importance} reports the influence values for the top 10 features (question–answers) in ClinStructor and all 8 sections in the MIMIC admission dataset using the Qwen3-8B CALM model. Features such as \textit{Admission Reason} and \textit{Age} are most influential for ClinStructor, while \textit{Present Illness} and \textit{Allergies} dominate in MIMIC3. 

\paragraph{Visualization at Individual Feature Levels: }

A key advantage of additive models such as NAMs and CALMs is the ability to directly visualize, for each feature $i$, the learned function $\ell_i(x_i) = F_{\text{last}}^i(F_T(X_i))$ across input values. This enables an intuitive understanding of how variations in a single feature affect the model’s output. Figure~\ref{fig:RiskTrends} illustrates this for four features from the ClinStructor Qwen3-8B model: \textit{Age}, \textit{Mental Status}, \textit{Respiratory Support}, and \textit{Admission Reason}. For each feature, we identify the 20 most frequent values, compute their risk scores $\ell_{i,n}^{(1)} - \ell_{i,n}^{(0)}$, and plot feature values against risk scores at the $0^{th}$, $25^{th}$, $50^{th}$, $75^{th}$, and $100^{th}$ percentiles.

These plots reveal clear patterns: risk increases with age; patients who are \textit{Unresponsive} have higher risk than those \textit{Awake and alert}; and admissions for \textit{Abdominal Pain} correlates more  higher risk than for \textit{Hematemesis} or \textit{Seizures}. Interestingly, they also reveal model's weaknesses, such as differing risk estimates for semantically equivalent inputs like \textit{Room air} and \textit{Room air (RA)}, indicating sensitivity to minor textual variations.

\subsection{Local Interpretability}
Local interpretability explains how a feature contribute to a model's prediction on a specific input.

\paragraph{Patient-Level Interpretability.}
For each patient, CALM provides feature-wise risk scores $\ell_{i,n}^{(1)} - \ell_{i,n}^{(0)}$. Figure~\ref{fig:patientinterpretability} illustrates this for two patients from the ClinStructor Qwen3-8B model—one who died and one discharged alive. For each patient, we plot the top five and bottom five contributing features based on risk scores. For Patient~1, the model assigns higher risk due to \textit{chronic conditions} and \textit{GI bleeding history}, and lower risk due to \textit{substance use} and \textit{heart rate}. For Patient~24, higher risk arises from \textit{malignancy history} and \textit{immunosuppressants}, while \textit{substance use} and \textit{neurologic deficits} contribute to lower risk.

\section{Conclusion}

In this work, we introduce \textbf{CALM}, an inherently interpretable framework for classification with semi-structured text. By decomposing the overall prediction into a sum of individual contributions, CALM provides faithful, component-level interpretation while maintaining performance comparable to black-box classifiers. Our extensions, CALM\textsuperscript{2} and CALM-Distill, further narrow this performance gap while preserving interpretability. Through evaluations using seven LLMs across three datasets, we demonstrate that CALM and its variants achieve competitive performance and enable transparent, model- and patient-level insights. Collectively, these results highlight CALM as a trustworthy solution for high-stakes domains.

\section{Limitations}
CALM enables us to interpret the model and understand how its predictions are made. However, it does not provide any mechanism to modify or correct the model’s behavior. It is important to note that the interpretations produced by CALM—such as feature importance—are specific to how the current model uses each feature, not to the inherent value or causal influence of the features themselves. For instance, if a particular feature strongly contributes to a high-risk prediction, this does not imply that the feature causes the high risk; rather, it indicates a strong correlation. As an example, if a certain medication is associated with higher predicted mortality risk, this does not mean the medication increases mortality, it may simply be prescribed for severe conditions that carry higher risk. In fact, the medication could even be protective.

\noindent\textbf{Ethical considerations}: 
Although our model is not designed for any specific application, it may be deployed in sensitive domains such as healthcare and related areas. We strongly recommend that any implementation of our method undergo thorough quality assurance and robustness evaluations prior to deployment in such settings. 

\noindent\textbf{License}: We use the PhysioNet datasets, which require credentialed access. All data and model usage in this work comply with their respective licenses.

\noindent\textbf{Replicability:}
All source code and the complete set of hyperparameters used in our experiments will be released publicly to facilitate transparency and reproducibility.

\bibliography{custom}

\newpage

\appendix
\section{Alternate Implementation of CALM (Packed Input)}

\subsection{Packed Concatenation with Per-Feature Isolation}

Let $X = \{X_1, \ldots, X_M\}$ with tokenizations $X_i = [x_{i1}, \ldots, x_{i|X_i|}]$.
Define the \emph{feature segment} for the $i$-th input as
\[
\mathrm{seg}(i) = X_i.
\]
The packed input stream is the concatenation of all segments:
\[
\text{Input} = \operatorname{concat}\big(\mathrm{seg}(i)\big)_{i=1}^{M},
\]
where $\operatorname{concat}(\cdot)$ denotes true sequence concatenation.

\paragraph{Block-Diagonal Attention Mask.}
Let $S_i$ be the set of token indices belonging to segment $\mathrm{seg}(i)$ within the packed stream.
The attention mask enforces isolation between features:
\[
\mathrm{AttnMask}(u,v) =
\begin{cases}
1, & \text{if } u,v \in S_i \text{ for some } i,\\[2pt]
0, & \text{otherwise.}
\end{cases}
\]
This ensures that tokens from different features cannot attend to each other.

\paragraph{Per-Feature Positional Re-Indexing.}
Let $\pi(t)$ denote the positional index at packed position $t$.
For any $u \in S_i$,
\[
\pi(u) = \mathrm{rank}_{S_i}(u) - 1,
\]
so \texttt{[START]}$_i$ has position $0$ and subsequent tokens are numbered sequentially within each segment, independent of their absolute position in the packed stream.

\paragraph{Feature Representations and Additive Logits.}
After the backbone forward pass:
\[
h_i = \mathrm{hidden}(\texttt{[EOS]}_i), \qquad
\ell_i = F_{\mathrm{last}}^{\,i}(h_i; W_{\mathrm{last}}^{\,i}) \in \mathbb{R}^C.
\]
CALM combines the feature logits additively:
\[
z = \frac{1}{M}\sum_{i=1}^M \ell_i + b, \qquad
p_c = \frac{\exp(z_c)}{\sum_{j=1}^C \exp(z_j)}.
\]

\paragraph{Equivalence to Direct CALM.}
Since attention is restricted within each $S_i$ and positional indices are reset per segment,
the backbone computes the same function for each $X_i$ as in independent forward passes.
Thus, reading $h_i$ at \texttt{[EOS]}$_i$ matches the direct implementation exactly.

\subsection{Time-Complexity Comparison}

Let $L_i = |X_i|$, $L_{\max} = \max_i L_i$, and $L_{\mathrm{tot}} = \sum_{i=1}^{M} L_i$.
We consider the quadratic self-attention cost, omitting constants and lower-order terms.

\paragraph{(1) Ideal CALM (Independent, No Padding).}
\[
\text{Cost} = \sum_{i=1}^{M} L_i^{2}.
\]
This represents the theoretical lower bound, where each feature is processed independently without padding.

\paragraph{(2) Practical CALM (Batched with Padding).}
\[
\text{Cost} = M \cdot L_{\max}^{2}.
\]
Here, all features in the batch are padded to the same maximum length $L_{\max}$,
leading to computational waste when feature lengths differ significantly.

\paragraph{(3) Packed Implementation.}
In the packed approach, all features are concatenated into a single sequence of total length $L_{\mathrm{tot}}$.
With a standard dense attention kernel, the cost is
\[
\text{Cost} = \left( \sum_{i=1}^M L_i \right)^{2}.
\]
This approach has two main properties:
\begin{itemize}
    \item It is always at least as large as the ideal cost $\sum_i L_i^2$,
    since cross-feature interactions are masked but still computed by the dense kernel.
    \item It can be smaller than the padded batch cost $M L_{\max}^2$ when feature lengths vary widely,
    as no computation is wasted on padding.
\end{itemize}
If a block-sparse attention kernel is used that exploits the block-diagonal mask,
the packed cost reduces exactly to the ideal $\sum_i L_i^2$.

\subsection{Compatibility with CALM2}

The packed implementation produces identical feature representations $h_i$
and is fully compatible with CALM$^2$.

\section{Comparison of CALM2 Vs. CALM: All metrics}

\begin{table*}[!t]
\centering
\begin{tabular}{llcccccc}
\toprule
 \multirow{2}{*}{Model} & \multirow{2}{*}{Rank, $\beta$}  &
\multicolumn{3}{c}{CALM$^2$} & \multicolumn{3}{c}{Finetune}  \\
\cmidrule(lr){3-5} \cmidrule(lr){6-8}
& & AUC-PR & F1 & AUC-ROC & AUC-PR & F1 & AUC-ROC \\
\toprule
\multirow{4}{*}{Qwen3-0.6B-Base}  & $Rank=8$, $\beta=0.1$    &   0.447 & 0.46 & 0.838 &  \multirow{4}{*}{0.457} & \multirow{4}{*}{0.471} & \multirow{4}{*}{0.85} \\
& $Rank=8$, $\beta=0.5$  & 0.445 & 0.453 & 0.841 & 
   \\
& $Rank=16$, $\beta=0.1$ &    0.437 & 0.462 & 0.843 & 
  \\
& $Rank=16$, $\beta=0.5$  & \textbf{0.464} & 0.467 & 0.845 & 
\\
\midrule
\multirow{4}{*}{Qwen3-1.7B-Base}  &  $Rank=8$, $\beta=0.1$   & 0.476 & 0.48 & 0.852 & \multirow{4}{*}{0.494} & \multirow{4}{*}{0.481} & \multirow{4}{*}{0.859} \\
& $Rank=8$, $\beta=0.5$  & \textbf{0.486} & 0.484 & 0.852 &   \\
& $Rank=16$, $\beta=0.1$ &   0.475 & 0.482 & 0.854 &   \\
& $Rank=16$, $\beta=0.5$  & 0.464 & 0.461 & 0.853 &   \\
\midrule
\multirow{4}{*}{Qwen3-8B-Base}  &  $Rank=8$, $\beta=0.1$   &   \textbf{0.504} & 0.49 & 0.864 &  0.532 & 0.506 & 0.874 \\
& $Rank=8$, $\beta=0.5$  &   0.482 & 0.49 & 0.856 & \\
& $Rank=16$, $\beta=0.1$ &    0.485 & 0.494 & 0.857 &    \\
& $Rank=16$, $\beta=0.5$  & 0.503 & 0.498 & 0.861 &\\
\midrule
\multirow{4}{*}{Phi-3.5-mini-instruct}  & $Rank=8$, $\beta=0.1$    &   \textbf{0.480} & 0.487 & 0.86 &  0.522 & 0.506 & 0.87 \\
& $Rank=8$, $\beta=0.5$  &  0.462 & 0.474 & 0.85 &    \\
& $Rank=16$, $\beta=0.1$ & 0.473 & 0.484 & 0.86 &  \\
& $Rank=16$, $\beta=0.5$  &   0.476 & 0.484 & 0.857 & \\
\midrule
\multirow{4}{*}{gemma-3-4b-pt}  &  $Rank=8$, $\beta=0.1$   &   0.487 & 0.487 & 0.859 &  0.51 & 0.49 & 0.868 \\
& $Rank=8$, $\beta=0.5$  &    0.5 & 0.494 & 0.863 & \\
& $Rank=16$, $\beta=0.1$ &    0.487 & 0.491 & 0.856 \\
& $Rank=16$, $\beta=0.5$  &   \textbf{0.492} & 0.491 & 0.857  \\
\bottomrule
\end{tabular}
\caption{ Comparison of CALM$^2$ Vs. CALM.}\label{tab:calm2appendix}
\end{table*}

\newpage

\section{Comparison of CALM-Distill Vs. CALM: All metrics}

\begin{table*}[!t]
\centering
\caption{ Comparison of CALM-Distill Vs. CALM.}\label{tab:distillappendix}
\begin{tabular}{llcccccc}
\toprule
 \multirow{2}{*}{Model} & \multirow{2}{*}{Temperature $\alpha$}  &
\multicolumn{3}{c}{CALM-Distill} & \multicolumn{3}{c}{Finetune}  \\
\cmidrule(lr){3-5} \cmidrule(lr){6-8}
& & AUC-PR & F1 & AUC-ROC & AUC-PR & F1 & AUC-ROC \\
\toprule
\multirow{3}{*}{Qwen3-0.6B-Base}  & 0.2 & 0.454 & 0.466 & 0.847 & \multirow{3}{*}{0.457} & \multirow{3}{*}{0.471} & \multirow{3}{*}{0.85} \\
& 0.4 & 0.445 & 0.463 & 0.848 &  \\
& 0.6 & 0.45 & 0.469 & 0.854 & \\
\midrule
\multirow{3}{*}{Qwen3-1.7B-Base}  &  0.2 & 0.479 & 0.472 & 0.857 & \multirow{4}{*}{0.494} & \multirow{4}{*}{0.481} & \multirow{4}{*}{0.859} \\
& 0.4 & 0.486 & 0.488 & 0.862 & \\
& 0.6 & 0.477 & 0.482 & 0.853 & \\
\midrule
\multirow{3}{*}{Qwen3-8B-Base}  & 0.2 & 0.512 & 0.508 & 0.868 &   0.532 & 0.506 & 0.874 \\
& 0.4 & 0.517 & 0.5 & 0.87 & \\
& 0.6 & 0.517 & 0.506 & 0.867 & \\
\midrule
\multirow{3}{*}{Phi-3.5-mini-instruct}  & 0.2 & 0.486 & 0.488 & 0.858 & 0.522 & 0.506 & 0.87 \\
& 0.4 & 0.495 & 0.488 & 0.865 & \\
& 0.6 & 0.503 & 0.505 & 0.868 & \\
\midrule
\multirow{3}{*}{gemma-3-4b-pt} &  0.2 & 0.505 & 0.494 & 0.864 & 0.51 & 0.49 & 0.868 \\ 
& 0.4 & 0.504 & 0.487 & 0.872 & \\
& 0.6 & 0.524 & 0.497 & 0.876 & \\
\bottomrule
\end{tabular}
\end{table*}

\section{Initialization of CALM model from Fully Finetuned Model}

\begin{table*}[htbp]
\centering
\begin{tabular}{lccc}
\toprule
\multirow{2}{*}{Model} &
\multicolumn{3}{c}{CALM$-FTInit$}  \\
\cmidrule(lr){2-4} 
& AUC-PR & F1 & AUC-ROC \\
\toprule
\multicolumn{4}{c}{Initialize Transformer Weights }\\
\toprule
 Qwen3-0.6B-Base   & 0.299 & 	0.329 & 	0.753    \\
 Qwen3-1.7B-Base   & 0.39 & 	0.401 & 	0.808 \\
 Qwen3-8B-Base     &0.408 & 	0.406 & 	0.815 \\
 \bottomrule
 \multicolumn{4}{c}{Initialize Transformer and Classifier Weights } \\
\bottomrule
Qwen3-0.6B-Base   & 0.273 & 	0.348 & 	0.74 \\
Qwen3-1.7B-Base   & 0.376 & 	0.379 & 	0.787\\
Qwen3-8B-Base     &  0.37 & 	0.392 & 	0.797    \\
\bottomrule
\end{tabular}
\caption{Comparison of Initialization from Fully Model: Initializing from Finetuned models makes it worse}
\end{table*}

\section{ClinStructor Questions ANd Influence Scores}

\begin{table*}[t!]
\centering
\small
\begin{tabular}{p{0.5cm}p{7cm}p{0.5cm}p{7cm}}
\toprule
\textbf{ID} & \textbf{Question} & \textbf{ID} & \textbf{Question} \\
\midrule
1 & what is the patient's age? & 2 & what chronic medical conditions does the patient have? \\
3 & what is the patient's blood pressure on admission? & 4 & what is the patient's heart rate on admission? \\
5 & what is the patient's reported history of tobacco, alcohol, or illicit drug use? & 6 & what is the patient's respiratory rate on admission? \\
7 & what is the primary reason for the patient's current admission? & 8 & what medications is the patient currently taking on admission? \\
9 & what is the patient's temperature on admission? & 10 & what is the patient's left ventricular ejection fraction? \\
11 & what is the patient's oxygen saturation on admission? & 12 & what is the patient's current mental status or level of consciousness? \\
13 & what type of respiratory support is the patient currently receiving? & 14 & what is the patient's gender? \\
15 & what is the patient's baseline creatinine level? & 16 & what chronic cardiovascular conditions does the patient have? \\
17 & what is the patient's history of malignancy, including type and treatment status? & 18 & what is the patient's living situation and social support structure? \\
19 & what antiplatelet or anticoagulant medications is the patient currently prescribed? & 20 & what is the patient's white blood cell count on admission? \\
21 & what is the patient's current oxygen saturation and the amount of supplemental oxygen required? & 22 & what are the patient's known drug allergies? \\
23 & what is the patient's history of hypertension? & 24 & what was the primary reason for the patient's intubation? \\
25 & what is the patient's history of diabetes mellitus, including type? & 26 & from what type of facility was the patient transferred? \\
27 & what are the key findings from the abdominal physical examination? & 28 & what is the patient's history of myocardial infarction? \\
29 & what is the reported severity of mitral regurgitation? & 30 & what chronic respiratory conditions does the patient have? \\
31 & what is the patient's body mass index (bmi)? & 32 & what significant family medical history is reported? \\
33 & what is the patient's inr on admission? & 34 & what specific neurological deficits are identified during the physical examination? \\
35 & how is the patient's general appearance described in the physical exam? & 36 & what is the patient's current renal replacement therapy status? \\
37 & what is the patient's employment status? & 38 & what immunosuppressive medications is the patient currently taking? \\
39 & what is the extent and location of any edema noted on physical examination? & 40 & what is the patient's lactate level? \\
41 & what are the key findings from the patient's respiratory physical examination? & 42 & what physical exam findings indicate the patient's hydration status? \\
43 & what was the mechanism of injury? & 44 & what was the patient's initial arterial blood gas ph? \\
45 & what is the patient's history of gastrointestinal bleeding? & 46 & what is the patient's history and current status regarding congestive heart failure? \\
47 & what is the patient's hematocrit level on admission? & 48 & what is the patient's primary cardiac diagnosis? \\
49 & what is the patient's troponin level? & 50 & what is the patient's history of coronary artery disease? \\
\bottomrule
\end{tabular}
\caption{50 Questions used in the ClinSructor Dataset}
\label{tab:appendix-questions}
\end{table*}

\begin{figure*}[!t]
    \centering
    \includegraphics[width=1\linewidth]{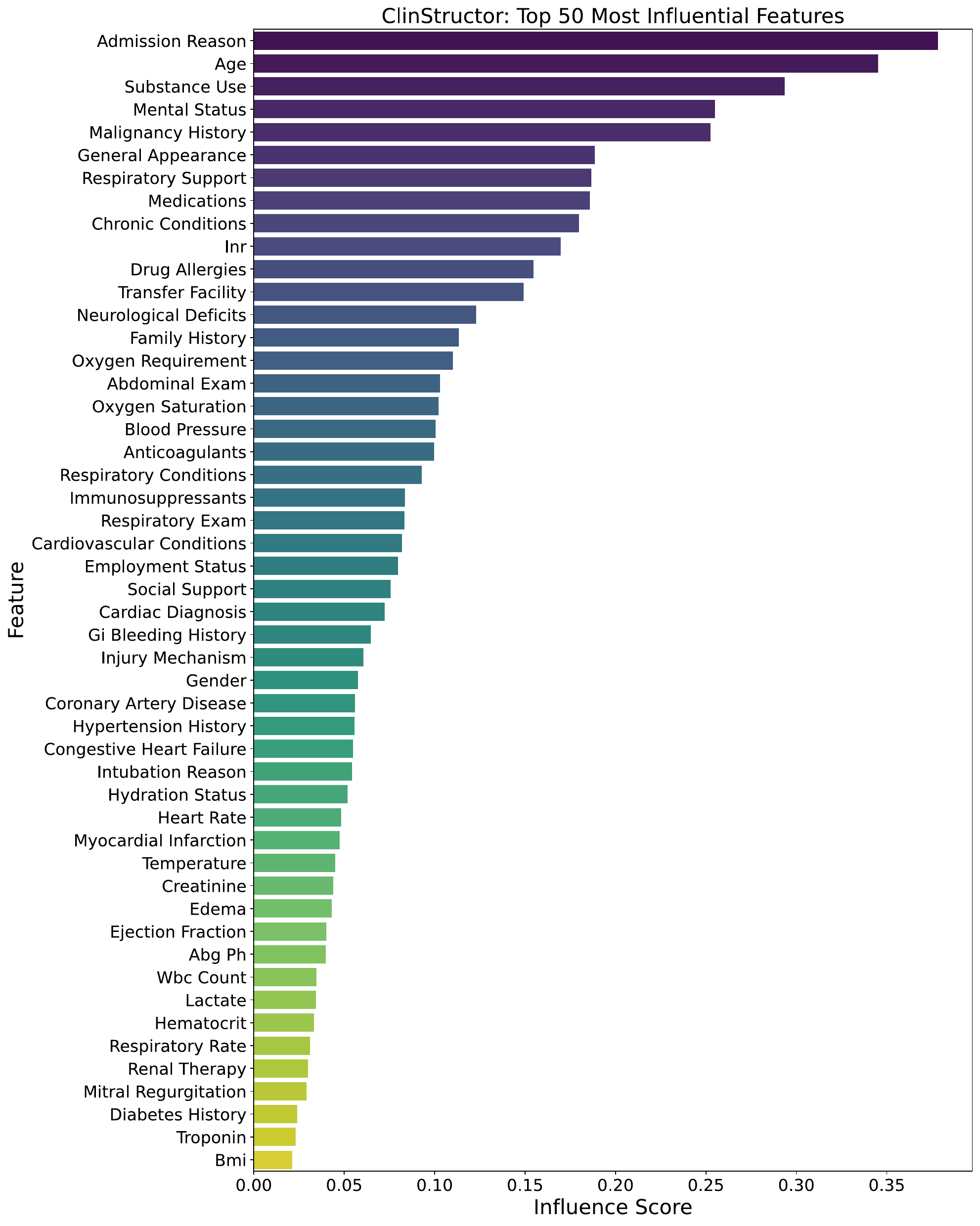}
    \caption{ClinStructor Influence Scores of ALl 50 features}
    \label{fig:placeholder}
\end{figure*}

\begin{figure*}[t!]
    \centering

    \begin{subfigure}[b]{0.24\textwidth}
        \centering
        \includegraphics[width=\linewidth]{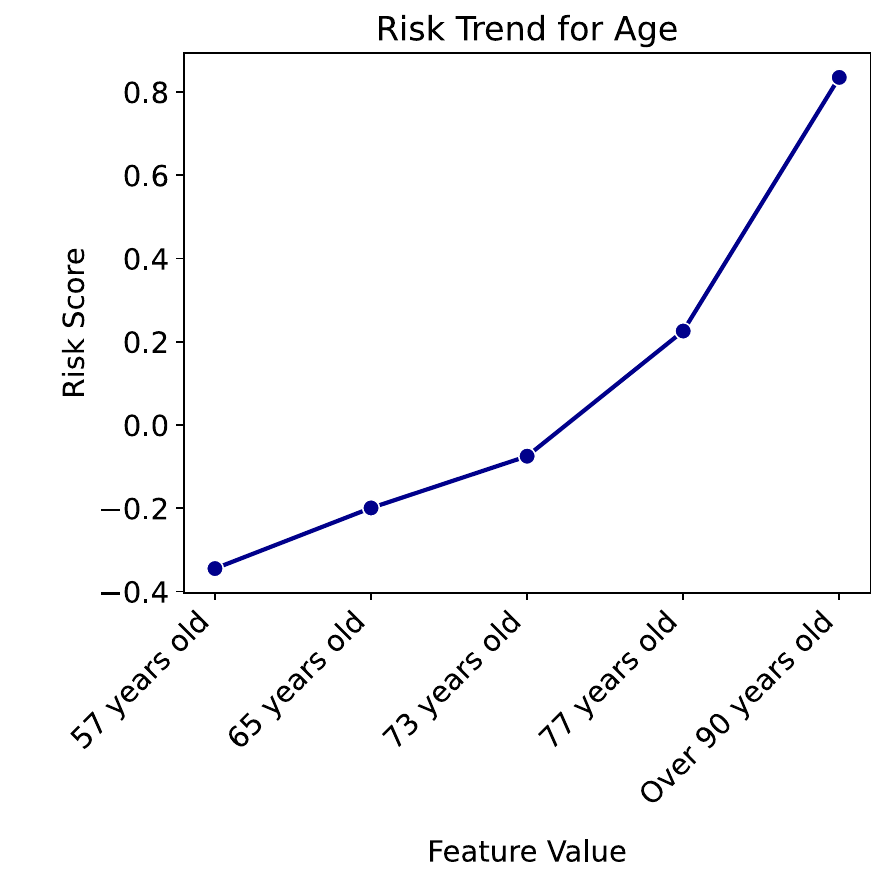}
        \caption{Feature 1}
    \end{subfigure}
    \begin{subfigure}[b]{0.24\textwidth}
        \centering
        \includegraphics[width=\linewidth]{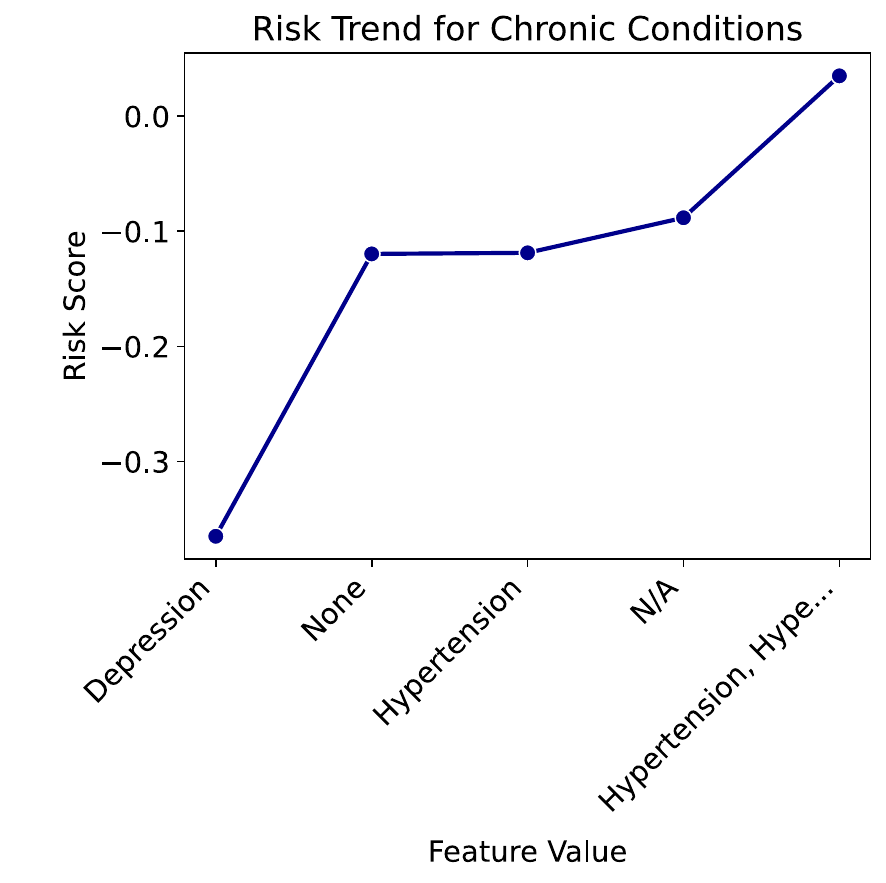}
        \caption{Feature 2}
    \end{subfigure}
    \begin{subfigure}[b]{0.24\textwidth}
        \centering
        \includegraphics[width=\linewidth]{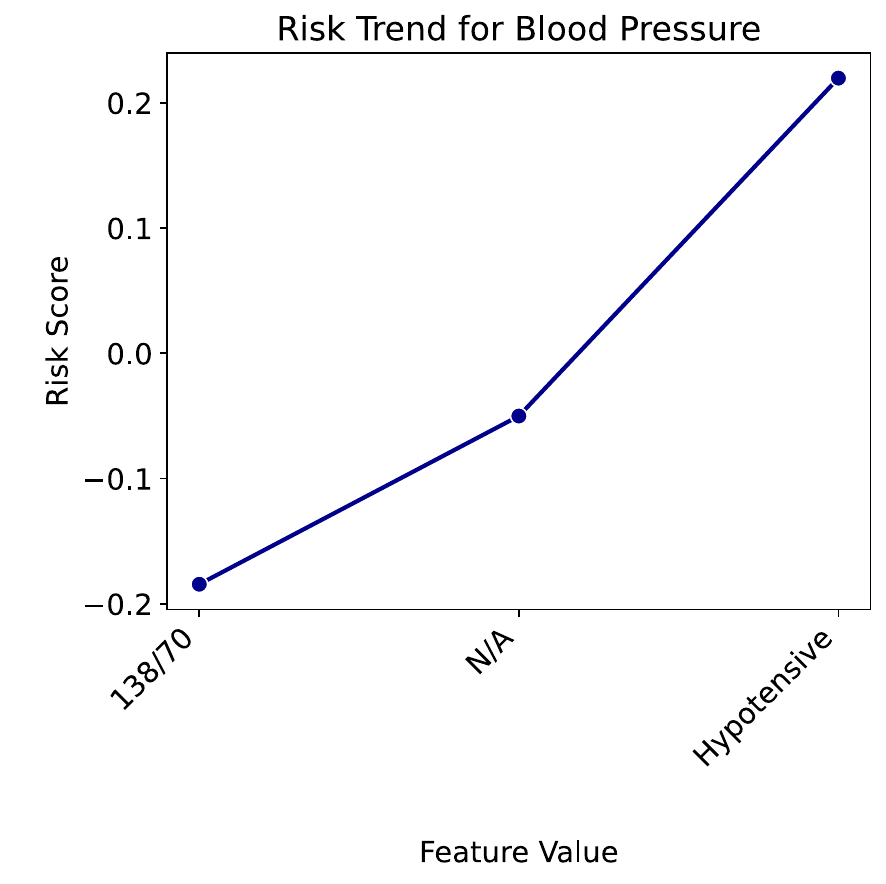}
        \caption{Feature 3}
    \end{subfigure}
    \begin{subfigure}[b]{0.24\textwidth}
        \centering
        \includegraphics[width=\linewidth]{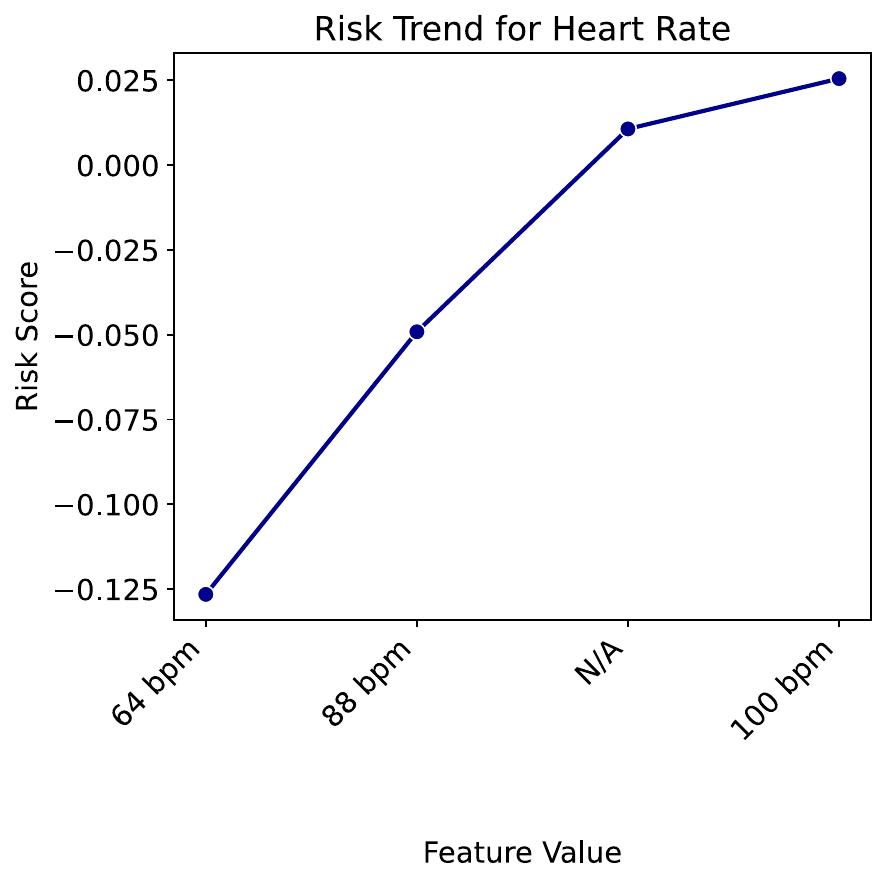}
        \caption{Feature 4}
    \end{subfigure}

    \par\vspace{0.8em}

    \begin{subfigure}[b]{0.24\textwidth}
        \centering
        \includegraphics[width=\linewidth]{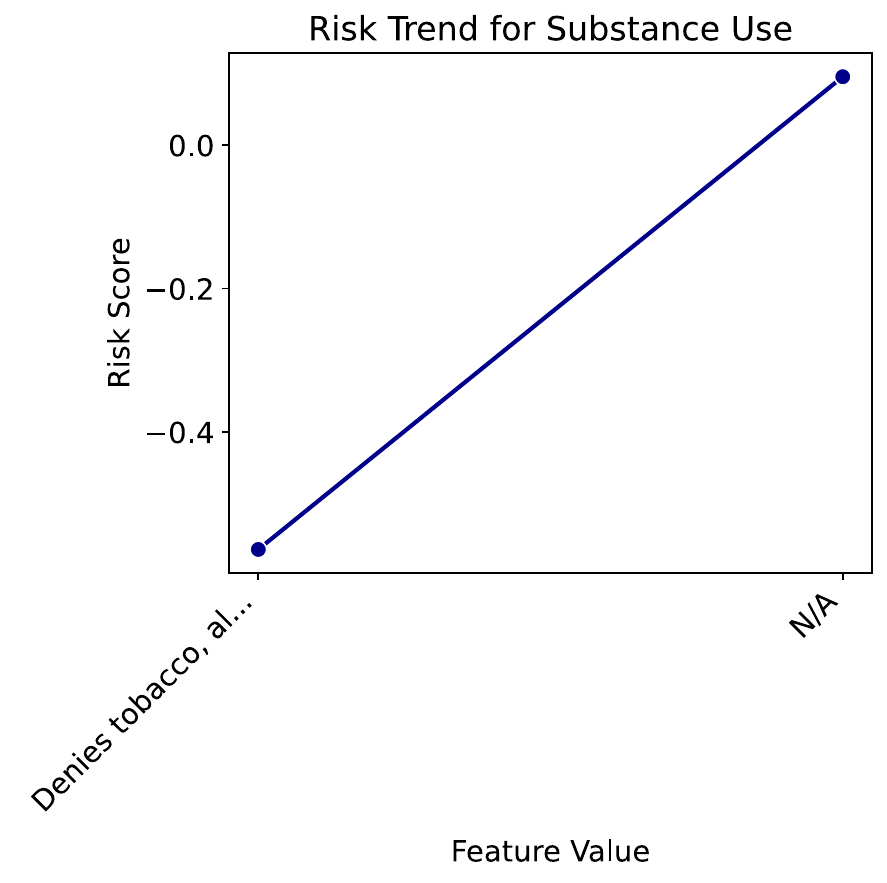}
        \caption{Feature 5}
    \end{subfigure}
    \begin{subfigure}[b]{0.24\textwidth}
        \centering
        \includegraphics[width=\linewidth]{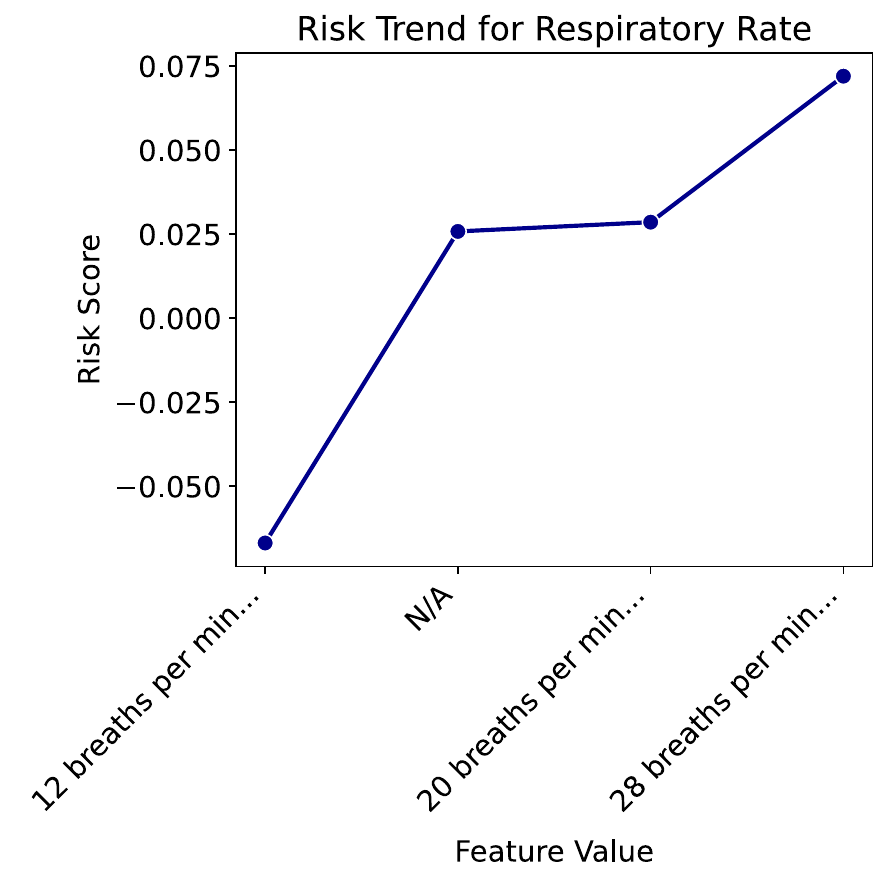}
        \caption{Feature 6}
    \end{subfigure}
    \begin{subfigure}[b]{0.24\textwidth}
        \centering
        \includegraphics[width=\linewidth]{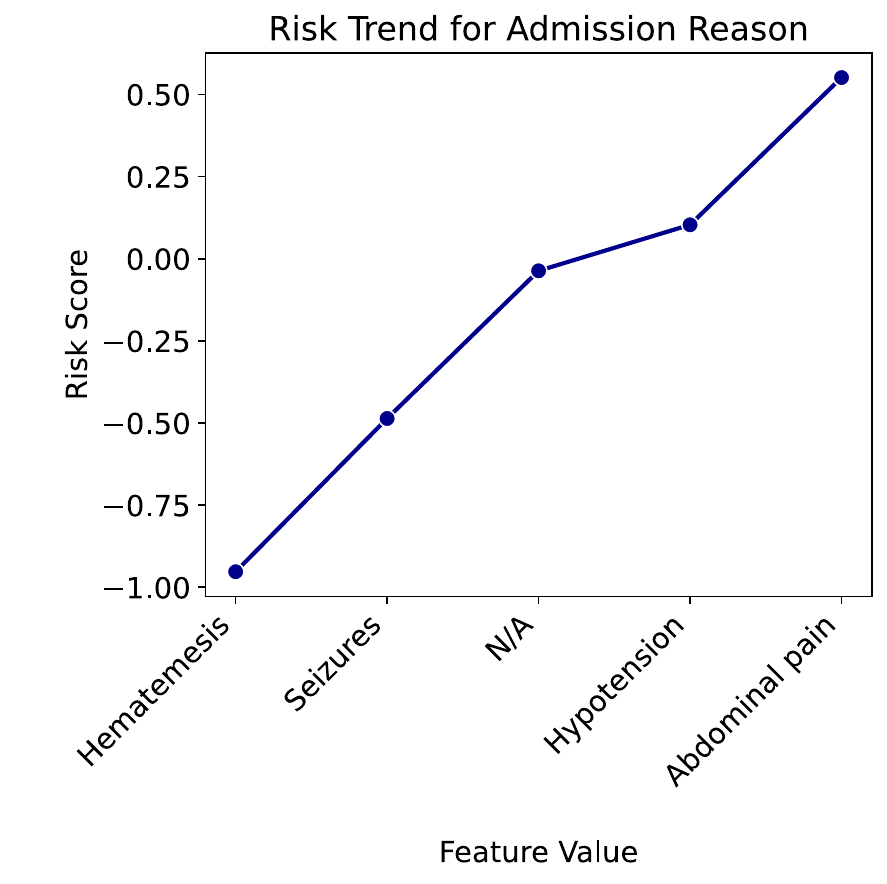}
        \caption{Feature 7}
    \end{subfigure}
    \begin{subfigure}[b]{0.24\textwidth}
        \centering
        \includegraphics[width=\linewidth]{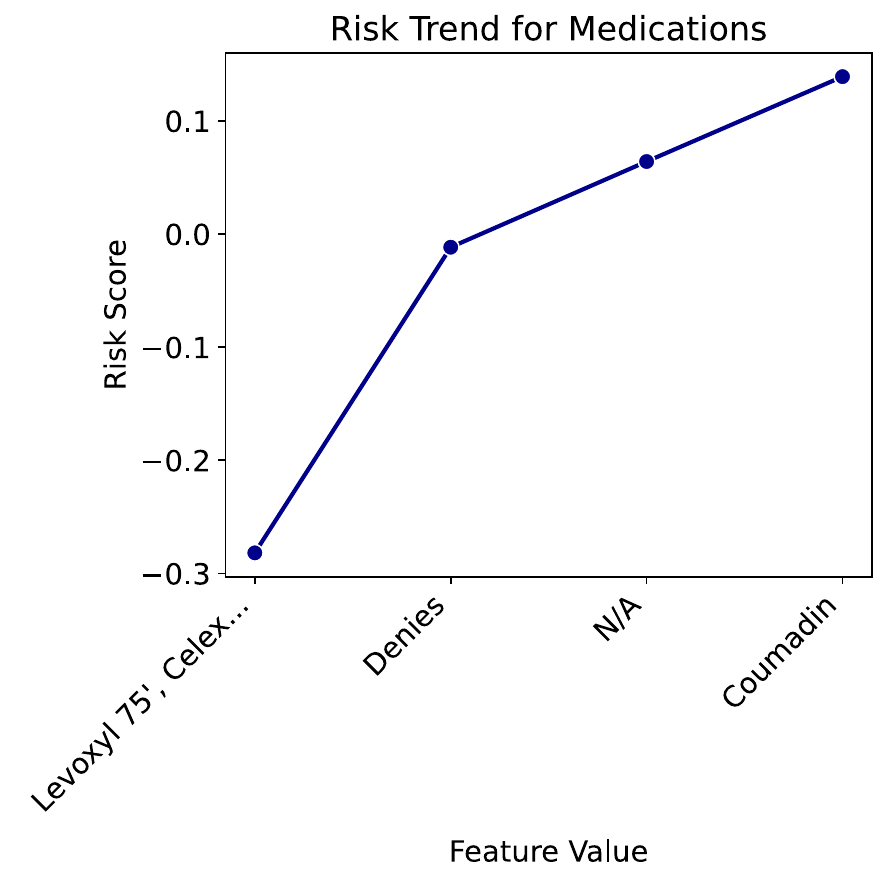}
        \caption{Feature 8}
    \end{subfigure}

    \par\vspace{0.8em}

    \begin{subfigure}[b]{0.24\textwidth}
        \centering
        \includegraphics[width=\linewidth]{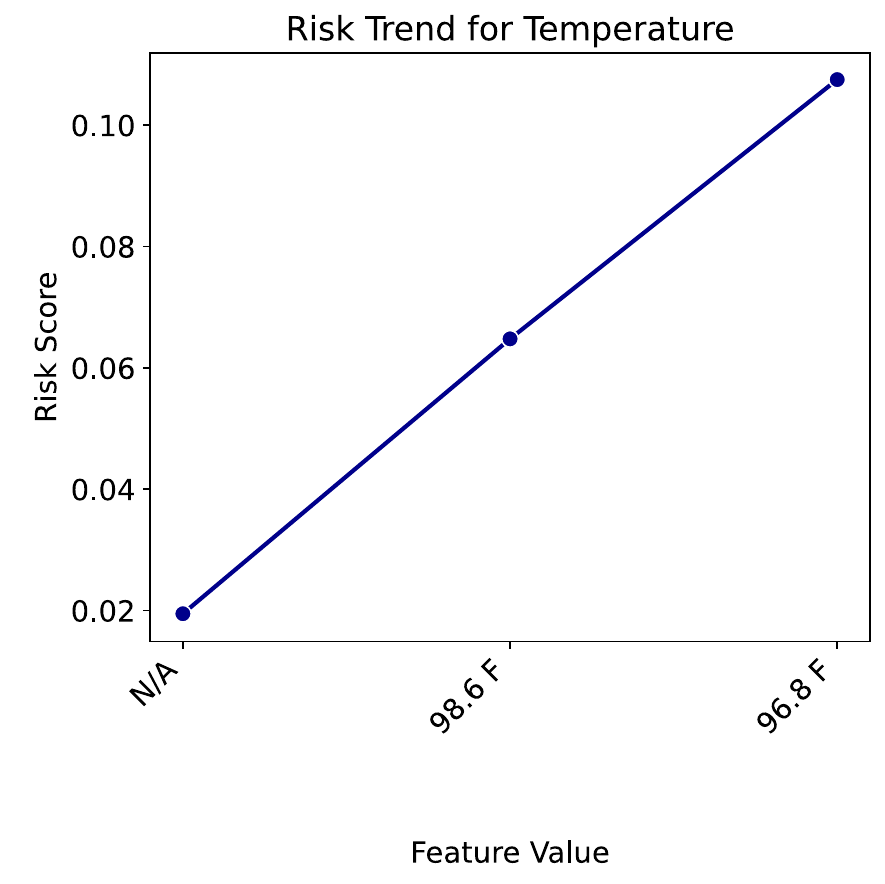}
        \caption{Feature 9}
    \end{subfigure}
    \begin{subfigure}[b]{0.24\textwidth}
        \centering
        \includegraphics[width=\linewidth]{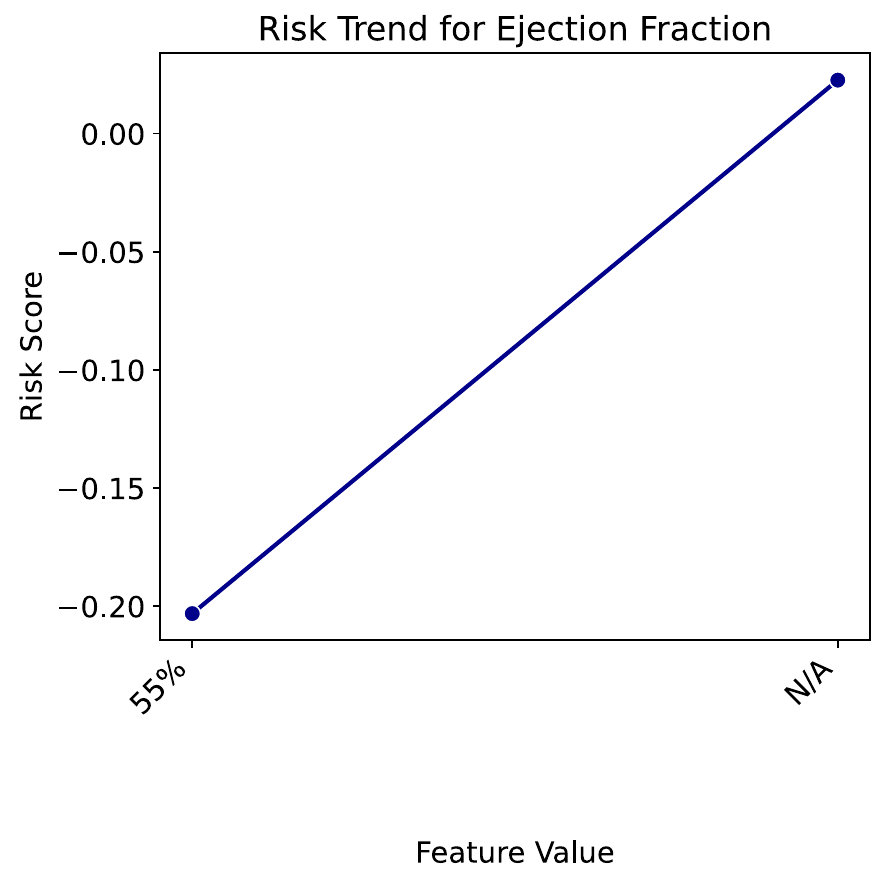}
        \caption{Feature 10}
    \end{subfigure}
    \begin{subfigure}[b]{0.24\textwidth}
        \centering
        \includegraphics[width=\linewidth]{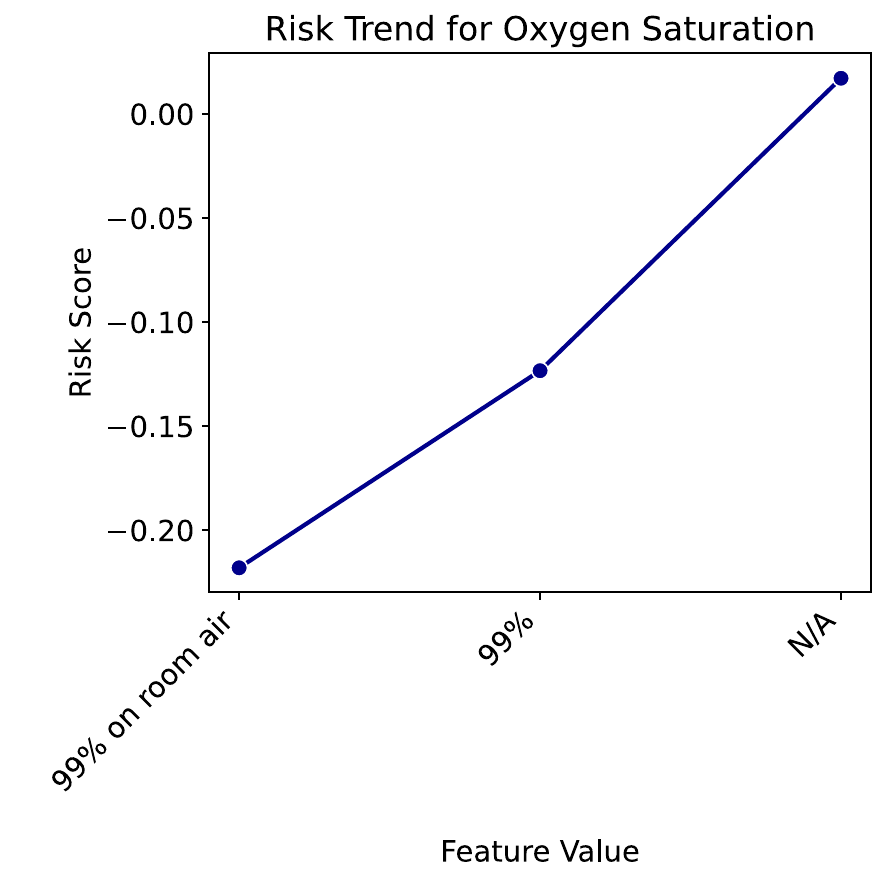}
        \caption{Feature 11}
    \end{subfigure}
    \begin{subfigure}[b]{0.24\textwidth}
        \centering
        \includegraphics[width=\linewidth]{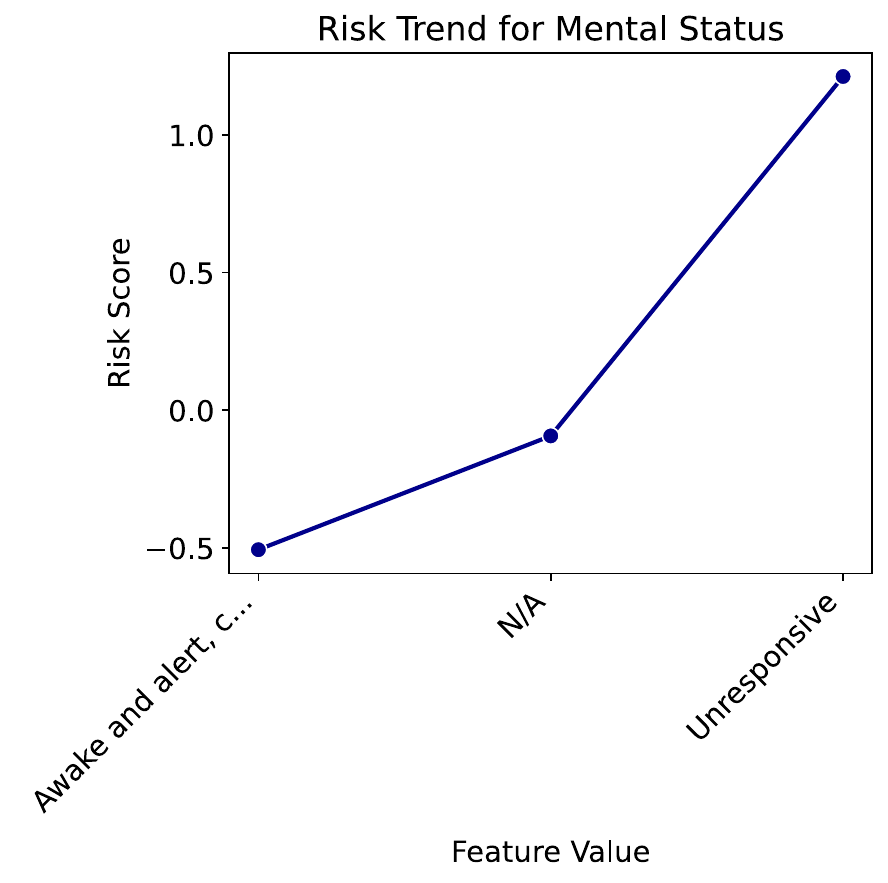}
        \caption{Feature 12}
    \end{subfigure}

    \par\vspace{0.8em}

    \begin{subfigure}[b]{0.24\textwidth}
        \centering
        \includegraphics[width=\linewidth]{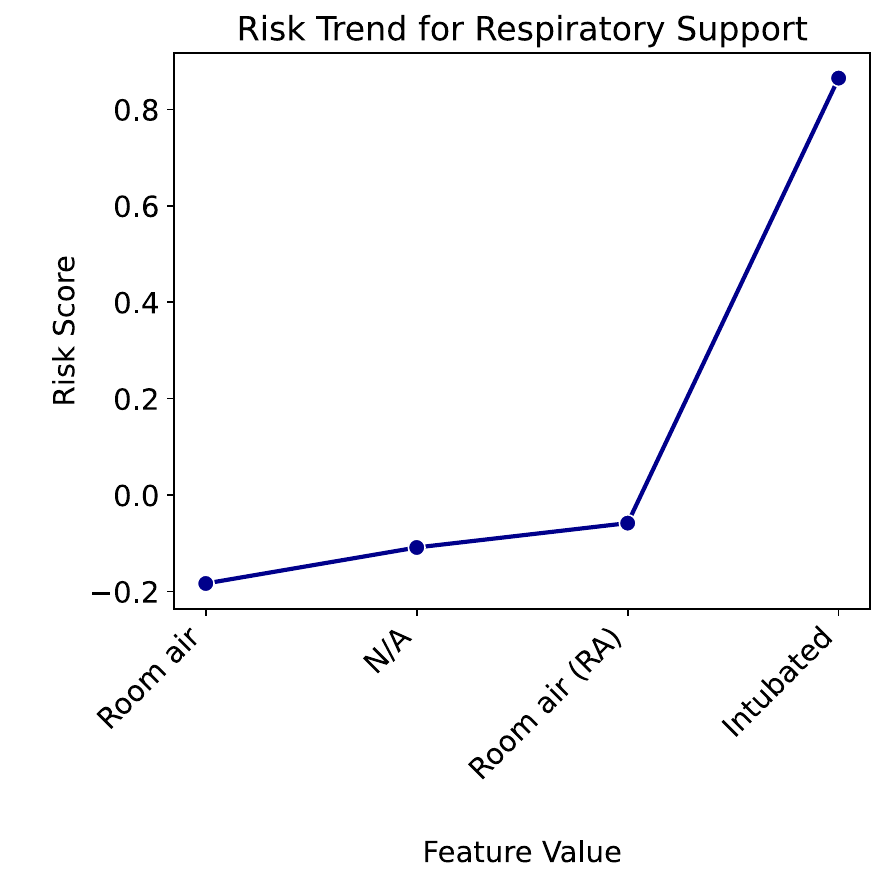}
        \caption{Feature 13}
    \end{subfigure}
    \begin{subfigure}[b]{0.24\textwidth}
        \centering
        \includegraphics[width=\linewidth]{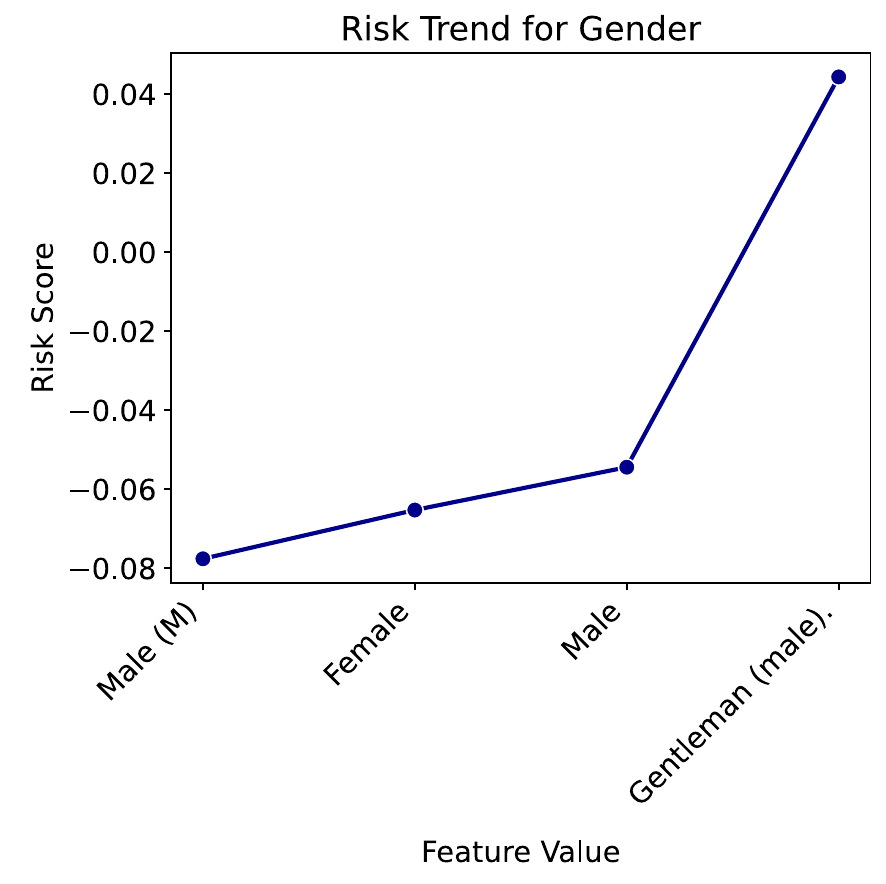}
        \caption{Feature 14}
    \end{subfigure}
    \begin{subfigure}[b]{0.24\textwidth}
        \centering
        \includegraphics[width=\linewidth]{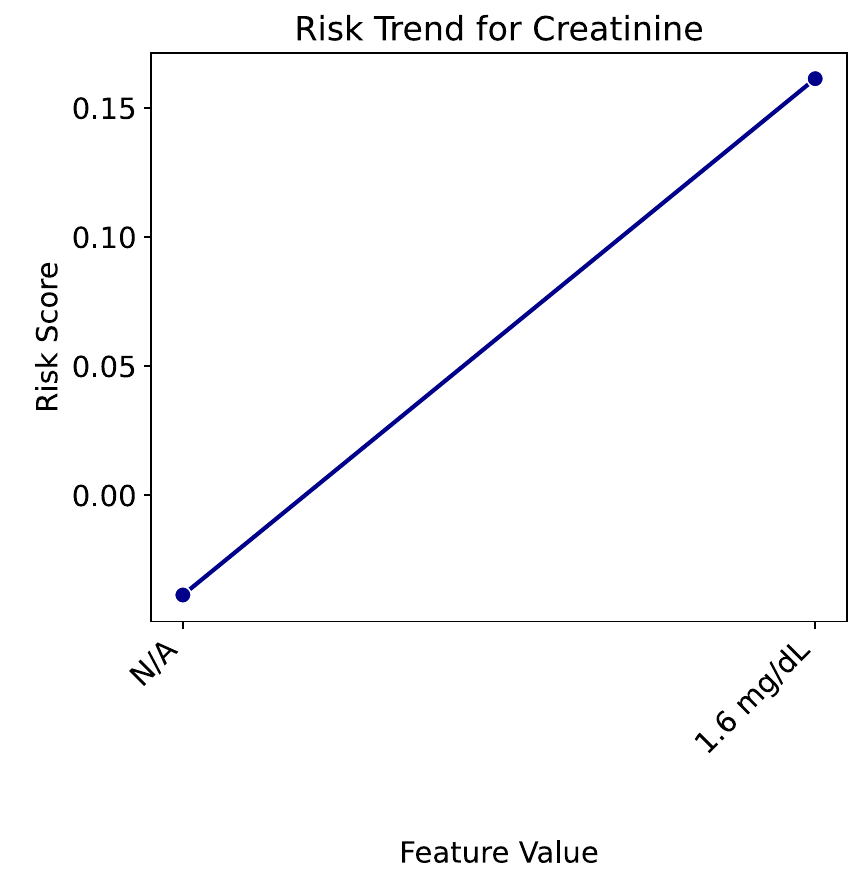}
        \caption{Feature 15}
    \end{subfigure}
    \begin{subfigure}[b]{0.24\textwidth}
        \centering
        \includegraphics[width=\linewidth]{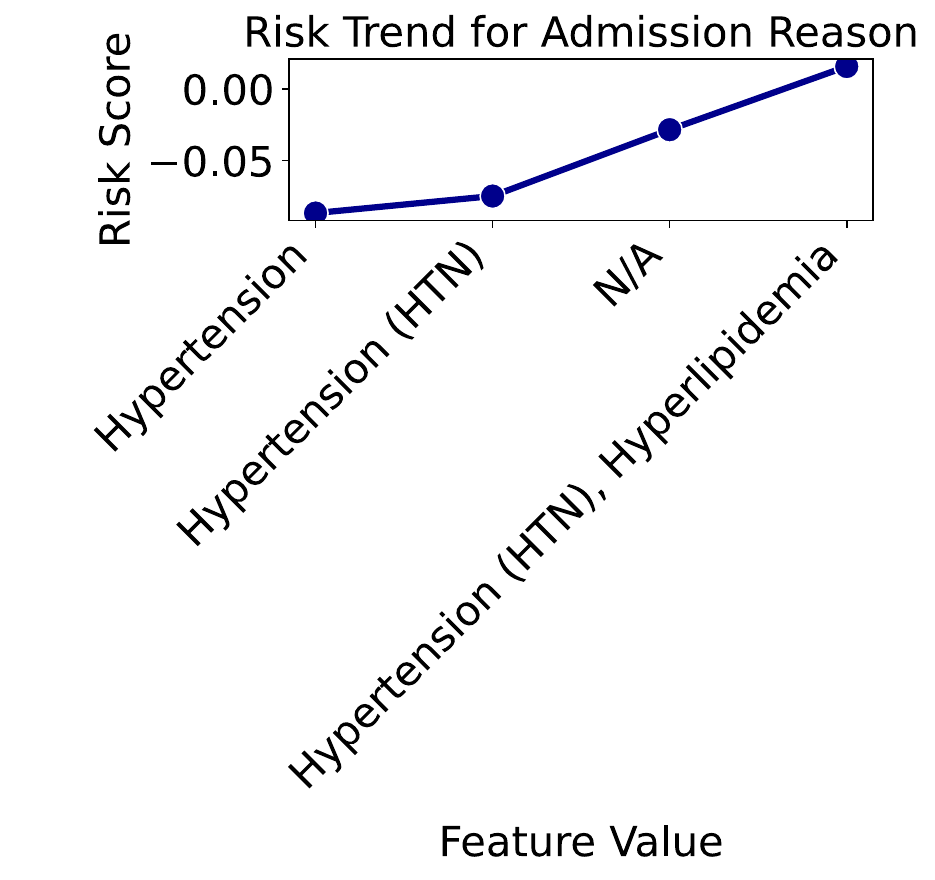}
        \caption{Feature 16}
    \end{subfigure}

    \par\vspace{0.8em}

    \begin{subfigure}[b]{0.24\textwidth}
        \centering
        \includegraphics[width=\linewidth]{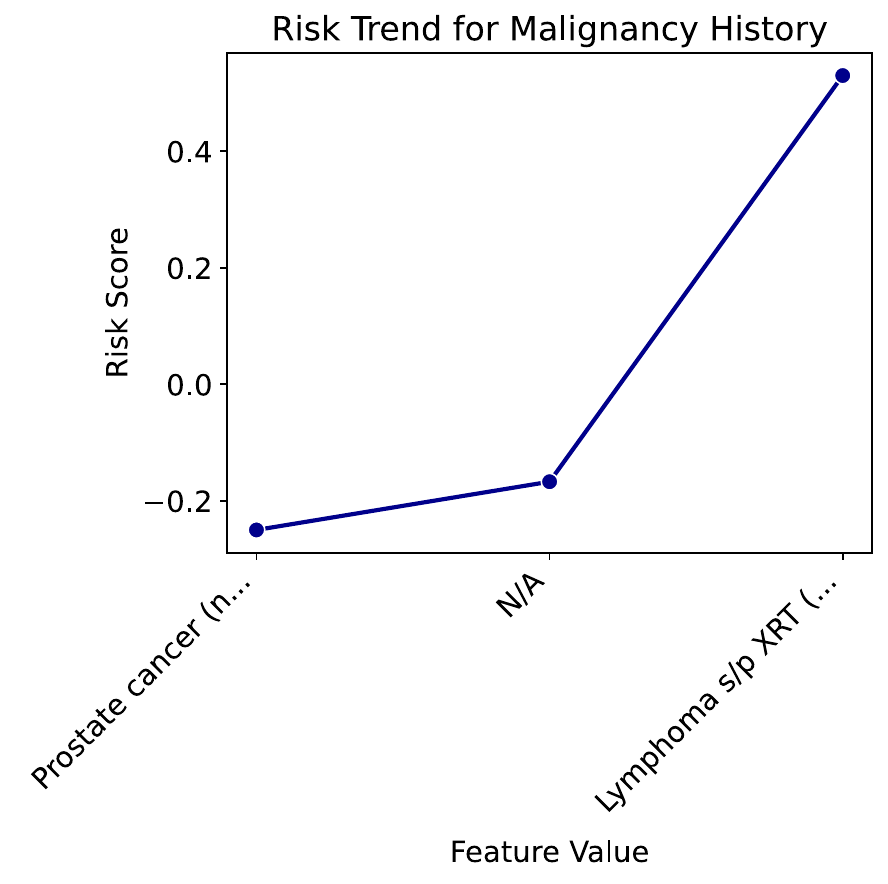}
        \caption{Feature 17}
    \end{subfigure}
    \begin{subfigure}[b]{0.24\textwidth}
        \centering
        \includegraphics[width=\linewidth]{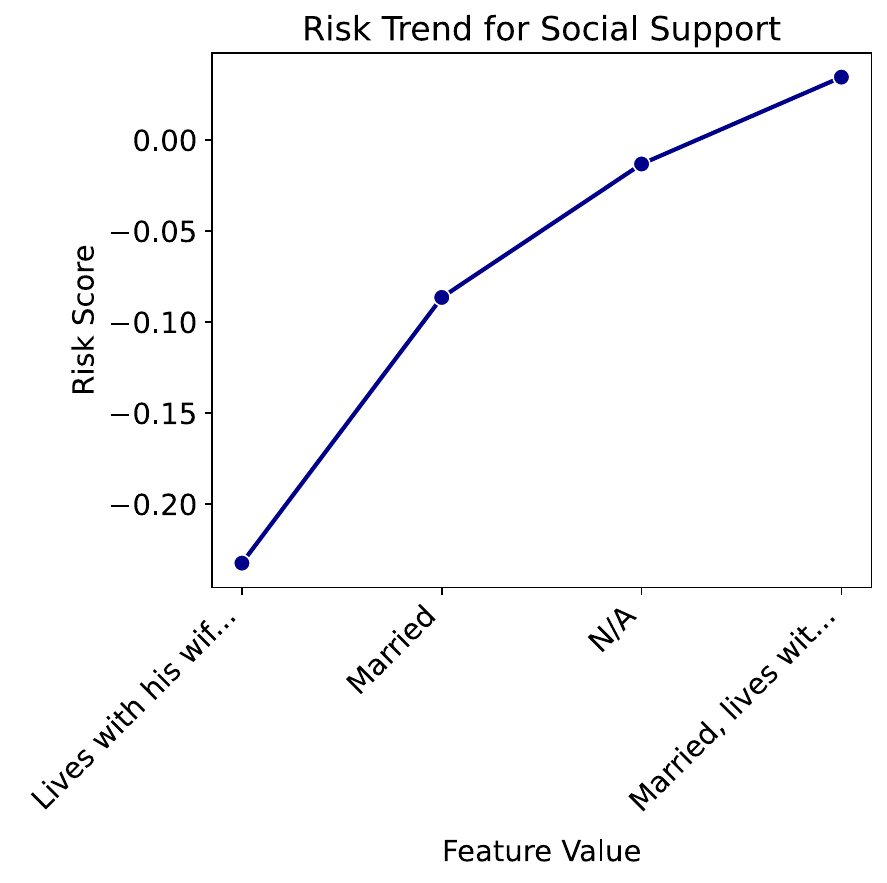}
        \caption{Feature 18}
    \end{subfigure}
    \begin{subfigure}[b]{0.24\textwidth}
        \centering
        \includegraphics[width=\linewidth]{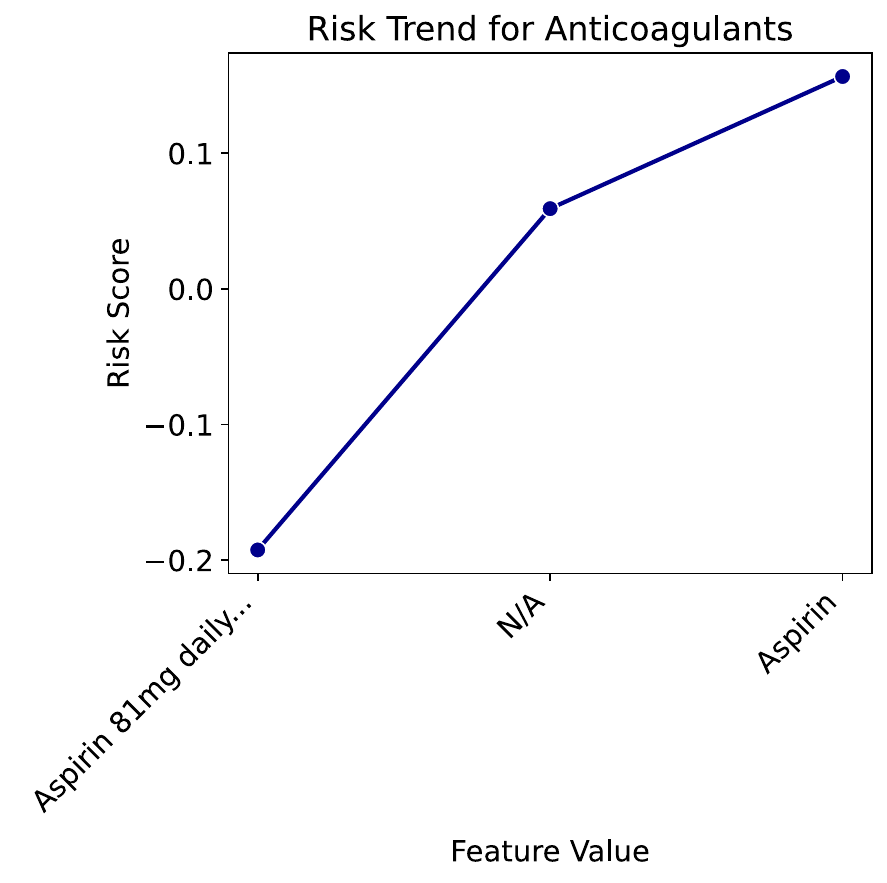}
        \caption{Feature 19}
    \end{subfigure}
    \begin{subfigure}[b]{0.24\textwidth}
        \centering
        \includegraphics[width=\linewidth]{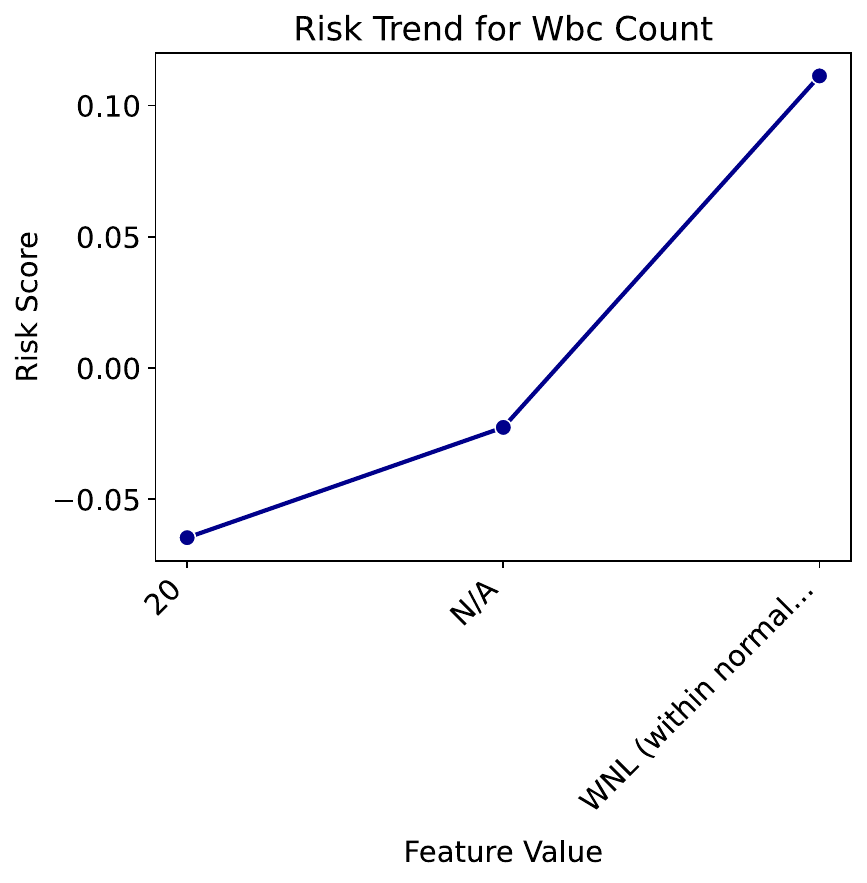}
        \caption{Feature 20}
    \end{subfigure}

    \caption{\textbf{Top 20 feature-level risk trends.}
    Each panel shows the learned logit function $\ell_i(x_i)$ for one of the top 20 most influential features, illustrating how model risk varies with feature value. The plots are ordered by overall importance.}
    \label{fig:Top20Features}
\end{figure*}

\section{Hyperparameter Settings}
\begin{verbatim}
configs = {
    "C1": {"lr": 1e-4, "rank": 8, "alpha": 16, "dropout": 0.05},
    "C2": {"lr": 1e-4, "rank": 8, "alpha": 32, "dropout": 0.05},
    "C3": {"lr": 1e-4, "rank": 16, "alpha": 16, "dropout": 0.05},
    "C4": {"lr": 1e-4, "rank": 16, "alpha": 32, "dropout": 0.05},
    "C5": {"lr": 2e-4, "rank": 8, "alpha": 16, "dropout": 0.05},
    "C6": {"lr": 2e-4, "rank": 16, "alpha": 32, "dropout": 0.05},
    "C7": {"lr": 5e-4, "rank": 8, "alpha": 32, "dropout": 0.05},
    "C8": {"lr": 5e-4, "rank": 16, "alpha": 16, "dropout": 0.05},
}
\end{verbatim}

\end{document}